\documentclass[sigconf, nonacm, natbib=false]{aamas}

\usepackage{balance} 

\usepackage{biblatex}
\usepackage{subcaption}
\usepackage{comment}
\usepackage{booktabs}
\usepackage{color, colortbl}
\usepackage{float}
\usepackage{amsmath}
\usepackage{graphicx}
\usepackage[table]{xcolor}
\definecolor{lightblue}{HTML}{E3EAF2}
\definecolor{lightgreen}{HTML}{e4f2e3}
\definecolor{lightred}{HTML}{e8c8c8}
\usepackage{bm}
\usepackage{makecell}
\usepackage{enumitem}
\usepackage{multirow}

\usepackage{algorithm}
\usepackage{algorithmic}
\usepackage{placeins}

\newcommand{\ENDIFadvice}[1]{\State \hlADVICE{\textbf{end if}}}

\newcommand{\ENDFORadvice}[1]{\State \hlADVICE{\textbf{end for}}}

\theoremstyle{plain}
\newtheorem{theorem}{Theorem}

\theoremstyle{definition}

\theoremstyle{remark}

\usepackage{tcolorbox}
\colorlet{myblue}{blue!40}

\setcopyright{ifaamas}
\acmConference[AAMAS '26]{Proc.\@ of the 25th International Conference
on Autonomous Agents and Multiagent Systems (AAMAS 2026)}{May 25 -- 29, 2026}
{Paphos, Cyprus}{C.~Amato, L.~Dennis, V.~Mascardi, J.~Thangarajah (eds.)}
\copyrightyear{2026}
\acmYear{2026}
\acmDOI{}
\acmPrice{}
\acmISBN{}

\acmSubmissionID{444}

\title[AAMAS-2026 Formatting Instructions]{Safe But Not Sorry: Reducing Over-Conservatism in Safety Critics via Uncertainty-Aware Modulation}

\author{Daniel Bethell}
\orcid{0000-0002-0685-5312}
\affiliation{
  \institution{University of York}
  \city{York}
  \country{UK}
}
\email{daniel.bethell@york.ac.uk}

\author{Simos Gerasimou}
\orcid{0000-0002-2706-5272}
\affiliation{%
  \institution{Cyprus University of Technology, Cyprus}
  \city{University of York}
  \country{York, UK}
}
\email{simos.gerasimou@cut.ac.cy}

\author{Radu Calinescu}
\orcid{0000-0002-2678-9260}
\affiliation{%
  \institution{University of York}
  \city{York}
  \country{UK}
}
\email{radu.calinescu@york.ac.uk}

\author{Calum Imrie}
\orcid{0009-0004-3198-9226}
\affiliation{%
  \institution{University of York}
  \city{York}
  \country{UK}
}
\email{calum.imrie@york.ac.uk}

\begin{abstract}
Ensuring the safe exploration of reinforcement learning (RL) agents is critical for deployment in real-world systems. Yet existing approaches struggle to strike the right balance: methods that tightly enforce safety often cripple task performance, while those that prioritise reward leave safety constraints frequently violated, producing diffuse cost landscapes that flatten gradients and stall policy improvement. We introduce the Uncertain Safety Critic (USC), a novel approach that integrates uncertainty-aware modulation and refinement into critic training. By concentrating conservatism in uncertain and costly regions while preserving sharp gradients in safe areas, USC enables policies to achieve effective reward–safety trade-offs. Extensive experiments show that USC reduces safety violations by $\approx 40\%$ while maintaining competitive or higher rewards, and reduces the error between predicted and true cost gradients by $\approx 83\%$, breaking the prevailing trade-off between safety and performance, and paving the way for scalable safe RL.
\end{abstract}


\newcommand{\BibTeX}{\rm B\kern-.05em{\sc i\kern-.025em b}\kern-.08em\TeX}

\begin{document}


\pagestyle{fancy}
\fancyhead{}


\maketitle 


\section{Introduction}
\label{sec:Introduction}

Reinforcement Learning (RL) empowers agents to autonomously devise sequential decision-making strategies by interacting with their environment~\cite{sutton2018reinforcement}. 
Its successes span a broad spectrum of domains, ranging from mastering strategic games~\cite{open-ai-five} to enabling sophisticated robotic control~\cite{loc-behaviours,massively-drl}, exhibiting capabilities on par with human cognition. 
Although exploration is the foundation through which an agent learns, it can become particularly problematic in safety-critical domains, such as healthcare and robotics, where physical or ethical risks may occur from unsafe actions.
During exploration in these domains, the agent can naturally encounter unfamiliar and potentially hazardous states in which erroneous actions can yield unacceptable consequences.  

%

Addressing this challenge mandates the design of RL agents that can both avoid unsafe behaviours and retain sufficient exploratory capacity to uncover optimal policies~\cite{concrete-problems}. Ensuring safety is  essential not only during training, when the agent explores and adapts, but also throughout deployment, where its decisions are observable in the real world and directly affect 
its stakeholders.

Prior safe RL approaches can be classified into external knowledge-based and cost-based.
External knowledge approaches leverage formal specifications~\cite {gross2022cool}, expert demonstrations~\cite{li2024guided}, or interactive oversight~\cite{saunders2017trial} to constrain exploration and prevent unsafe behavior~\cite{ odriozola2023shielded}.
Despite their effectiveness, they often require substantial domain expertise or supervision~\cite{bastani2018verifiable}. 
Cost-based methods, instead, embed safety directly into the learning problem by formulating RL as a constrained optimisation task within the Constrained MDP (CMDP) framework~\cite{altman1998constrained}. 
Within this scheme Lagrangian relaxation~\cite{stooke2020responsive}, primal–dual optimisation~\cite{paternain2022safe}, and refined objectives~\cite{satija2020constrained,yu2019convergent} balance reward against cumulative cost. 
These methods, however, rely on estimates of expected future cost produced by safety critics, whose estimation capability is severely affected by noisy or sparse cost signals~\cite{srinivasan2020learning}.
In contrast, conservative variants assume that uncertain or underexplored states might incur higher risk than in reality and mitigate underestimation by biasing toward higher costs~\cite{bharadhwaj2020conservative}.

Safety critics~\cite{bharadhwaj2020conservative,srinivasan2020learning}, while central to enforcing safety constraints, often produce distorted gradient signals that hinder learning~\cite{castellano2023learning}. By uniformly inflating or flattening cost estimates across large regions of the state–action space, these critics blur the boundary between genuinely unsafe and merely uncertain actions. 
Consequently, the gradients passed to the policy become weak, saturated, or misaligned with the underlying risk landscape. In already safe regions, inflated cost signals can suppress reward gradients, preventing policy refinement and improvement. Conversely, when the dual variable in the Lagrangian framework is high, the critic’s outputs may saturate, marking nearly all actions as unsafe and collapsing the gradient structure~\cite{castellano2023learning}. This collapse leaves the policy without informative guidance, stalling progress and undermining the delicate trade-offs between reward maximisation (performance) and constraint satisfaction (safety).

We address this significant limitation by introducing \textbf{U}ncertain \textbf{S}afety \textbf{C}ritic (USC), a novel approach that reduces over-conservatism in safety critics by explicitly considering parameter-space uncertainty in cost estimation. 
Unlike standard safety critics~\cite{srinivasan2020learning, bharadhwaj2020conservative}, which often flatten or inflate risk landscapes indiscriminately, USC selectively modulates conservatism proportionally to the critic’s epistemic uncertainty, preserving informative gradients while still erring on the side of caution. Drawing inspiration from recent advances in influence-based uncertainty quantification~\cite{koh2017understanding}, USC shapes its conservative bias through uncertainty-weighted objectives and further sharpens predictions via an uncertainty refinement procedure that interpolates poorly represented regions of the state–action space. By doing so, USC produces sharper and more reliable cost maps, enabling policies to achieve scalable and precise reward–cost trade-offs, with gradient errors reduced by $\approx 83\%$ compared to the state-of-the-art, a result that we further substantiate through our theoretical analysis. Moreover, USC reduces safety violations by up to $40\%$ while simultaneously achieving higher rewards that scale effectively with task complexity across nearly all evaluated settings. To the best of our knowledge, USC is the first safety critic method that couples conservative estimation with uncertainty-aware modulation, directly addressing the gradient collapse that undermines existing RL safety critics.


\section{Related Work}
\label{sec:Related Work}


\textbf{Approaches Based on External Knowledge.} Several  strategies for enabling safe exploration in reinforcement learning leverage prior knowledge or explicit safety constraints~\cite{gu2024review}. By extracting the agent's policy during learning and utilising formal verification techniques, these policies can be verified against predefined safety specifications, providing guarantees on their behaviour before deployment~\cite{bastani2018verifiable, gross2022cool}.
An analogous way to utilise prior knowledge for safe exploration involves using expert demonstrations to initialise the agent’s understanding of task dynamics and safety-critical behaviour~\cite{li2024guided, ramirez2022model}. Similarly, interactive oversight during training allows experts to intervene when the agent is likely to take a dangerous action~\cite{xu2022look, saunders2017trial, plaut2025asking}, or to design a curriculum that gradually exposes the agent to more challenging or risky scenarios~\cite{turchetta2020safe}.
An alternative approach employs safety shields, which intercept or correct unsafe actions during execution~\cite{konighofer2020shield,odriozola2023shielded}. While some shields rely on predefined safety constraints through LTL~\cite{alshiekh2018safe} or probabilistic specifications~\cite{jansen2020safe}, this requires domain knowledge; others can be learned online~\cite{shperberg2022learning}, partially relaxing the need for prior knowledge.
While these techniques can be highly effective, they rely to varying degree on often unavailable or costly expert input, domain knowledge, or structured priors. This motivates an alternative direction: enabling safety via environmental feedback alone, without assuming external supervision or formal task knowledge~\cite{bethell2024safe}.


\smallskip\noindent
\textbf{Cost-based Learning Approaches.} In this setting, safety is no longer enforced through external supervision or logical constraints, but instead learned through interaction with the environment~\cite{wachi2024survey}. By incorporating cost signals into the MDP framework to provide the agent with feedback on risky actions or outcomes~\cite[p.~21]{altman2021constrained}, similar to task feedback via reward, the agent can be trained to balance task performance against cumulative risk, without requiring prior knowledge of what constitutes safe behaviour. A common approach is to formulate the agent’s objective as a constrained optimisation problem, where the goal is to maximise expected reward subject to an upper bound on expected cumulative cost. Dual-based methods that use a Lagrangian relaxation of the constraint are among the most widely adopted approaches in this setting. These methods introduce a cost multiplier to penalise constraint violations during policy learning, effectively transforming the constrained problem into unconstrained by augmenting the reward signal~\cite{altman1998constrained, stooke2020responsive}. However, this formulation can lead to instability or poor convergence if the multiplier is not well-tuned or if cost gradients are unreliable. Primal-dual optimisation~\cite{paternain2022safe} jointly updates the policy and dual variables using structured learning rules to guarantee convergence. RCPO~\cite{tessler2018reward} decouples the reward and cost learning signals to avoid interference, using a soft constraint formulation that adjusts the constraint budget based on policy performance. More recently, methods such as Convergent Policy Optimization~\cite{yu2019convergent} and Backwards Value Functions~\cite{satija2020constrained} have introduced reformulations of the CMDP objective that ensure convergence to feasible policies, even under complex constraint landscapes or sparse cost signals. Across all these methods, however, a shared and often under-examined challenge remains: accurately estimating expected cumulative cost. Since policies are trained to balance reward against predicted cost, the reliability and structure of the cost signal, typically learned through a safety critic, play a critical role in both constraint satisfaction and overall performance.

\begin{figure}[tb]
    \centering
    \includegraphics[width=0.9\linewidth]{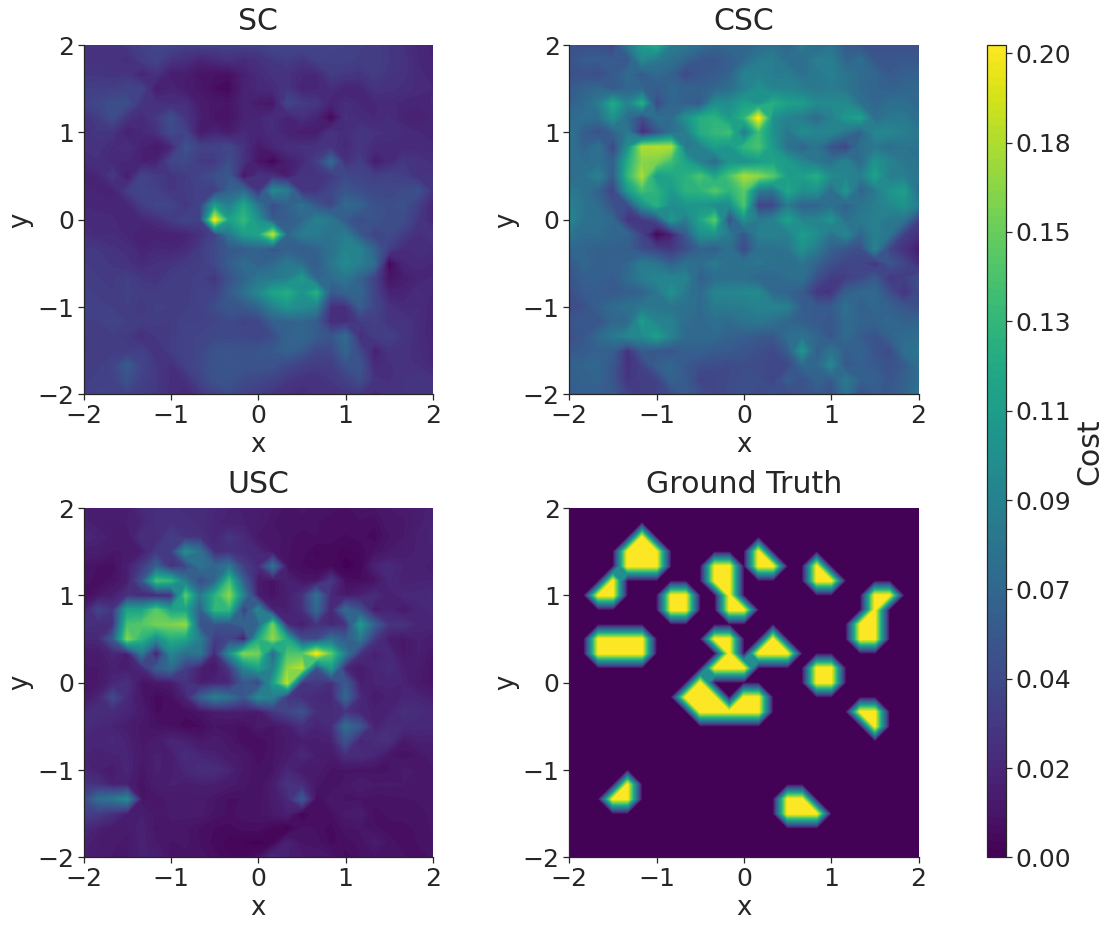}
    \vspace{-4mm}
    \caption{A qualitative comparison of the predictive cost maps from the SC, CSC, and USC methods in the CarGoal2 environments against the ground truth cost map.}
    \label{fig:cargoal2-costmap}
    \Description[cost map]{cost map}
    \vspace{-4mm}
\end{figure}

\smallskip\noindent
\textbf{Safety Critic (SC) Approaches.} Central to the success of cost-based approaches is the accurate estimation of expected cumulative cost, which informs both the policy’s constraint satisfaction and its trade-off with reward~\cite{gu2024review}. This estimation is typically carried out by a safety critic, a value function trained to predict future cost under the current policy~\cite{srinivasan2020learning}. In practice, standard critics often struggle with sparse or high-variance cost signals, leading to unstable or inaccurate learning. To mitigate this, recent research has proposed conservative safety critics (CSC)~\cite{bharadhwaj2020conservative} that deliberately overestimate cost across the state-action space. Rather than aiming to estimate the true expected cost, CSC encourages the critic to assign higher values by penalising low-cost predictions for new policy actions. While this approach can reduce the risk of underestimating safety violations, it often produces diffused or inflated cost landscapes that obscure genuine risk boundaries. Consequently, policies trained under CSC may receive weak or saturated gradients, limiting their ability to make fine-grained reward–cost trade-offs. For example, Figure~\ref{fig:cargoal2-costmap} illustrates the predictive cost maps generated by the different safety critic variants in the CarGoal2 task of the Safety Gymnasium test-suite~\cite{ji2023safety}. While the SC underestimates hazard boundaries, the CSC agent inflates cost uniformly across the environment space, producing diffuse cost maps that obscure genuine risk regions.


\section{Preliminaries}
\label{sec:Preliminaries}
\textbf{Markov Decision Process.}  
A Markov Decision Process (MDP)~\cite{bellman1957markovian} is a discrete-time stochastic control framework for sequential decision-making. 
An MDP is defined as the tuple $ M = (S, A, P, R, \gamma) $, where $ S $ is the state space, 
$ A $ is the action space, $ P(s_{t+1} | s_t, a_t) $ denotes the probability of transitioning to state $ s_{t+1} $ from state $ s_t $ after taking action $ a_t $, $ R(s,a) $ is the reward function, and $ \gamma \in [0,1) $ is the discount factor. A policy $ \pi: S \to \Delta(A) $ maps states to distributions over actions. Classical solution methods like value and policy iteration~\cite{bertsekas2008introduction} assume full knowledge of the MDP dynamics. MDPs include \emph{accepting states}, which represent task completion, and \emph{non-accepting terminal states}, which often signal safety violations. In this work, we consider MDPs with continuous state and action spaces, i.e., $S \subseteq \mathbb{R}^n $, $ A \subseteq \mathbb{R}^m $.

A Constrained Markov Decision Process (CMDP)~\cite{altman2021constrained} extends the standard MDP framework to incorporate safety or resource constraints. Formally, a CMDP is defined as a tuple $ M' = $($S$, $A$, $P$, $R$, $C$, $\chi$, $\gamma$) where ($S$, $A$, $P$, $R$, $\gamma$) defines the underlying MDP, $C: S \times A \to \mathbb{R}_{\geq 0}$ is a cost function that assigns a non-negative cost to each state-action pair, and $\chi \in \mathbb{R}_{\geq 0}$ denotes the maximum allowable cost per episode. The objective is to find a policy $\pi^*$ that maximises the expected return $\mathbb{E}\left[\sum_{t=0}^\infty \gamma^t R(s_t, a_t)\right]$
while satisfying the constraint $\mathbb{E}\left[\sum_{t=0}^\infty \gamma^t C(s_t, a_t)\right] \leq \chi$. CMDPs provide a principled approach to encode safety or performance limits directly into the learning process and are commonly addressed using Lagrangian-based methods or constrained optimisation techniques. In this work, CMDPs are used as a foundation for modelling safe decision-making under soft safety constraints, which are enforced in expectation and become hard when the safety budget $\chi=0$.

\textbf{Reinforcement Learning.}  
Reinforcement Learning (RL) addresses decision-making in unknown environments modelled as MDPs. The agent seeks an optimal policy $ \pi^* $ that maximises the expected discounted return $ \mathbb{E}\left[\sum_{t=0}^{\infty} \gamma^{t} R(s_t, a_t)\right] $~\cite{sutton2018reinforcement}. The value function $ V_\pi(s) $ estimates the expected return from state $ s $ under policy $ \pi $. Regret~\cite{berry1985bandit} quantifies the performance gap between the learned and optimal policies, with typical rates $ O(1/\sqrt{E_\mathit{max}}) $, where $ E_\mathit{max} $ is the training horizon. Canonical algorithms include Q-learning~\cite{qlearning} and SARSA~\cite{sarsa}. Deep Reinforcement Learning (DRL) employs neural networks to approximate policies or value functions in high-dimensional settings. Actor-critic methods~\cite{sutton2018reinforcement} decouple the policy (actor) and value function (critic), enabling concurrent updates. The Deep Deterministic Policy Gradient (DDPG) algorithm~\cite{ddpg-paper} exemplifies this approach in continuous action domains.

\textbf{Safety Critics.}
A safety critic is a learned function $Q_C(s, a)$ estimating the cost of executing action $a$ in state $s$. In constrained RL, such critics are commonly used within Lagrangian formulations to enforce expected cost constraints by guiding updates to a dual variable $\lambda$ that penalises unsafe behaviour. They are particularly important when only sparse or binary signals indicate constraint violations. In Conservative Safety Critics (CSC), $Q_C$ is explicitly trained to overestimate risk, encouraging cautious exploration. However, this design introduces several failure modes. First, conservative critics often produce flattened cost maps, assigning uniformly high values that obscure distinctions between dangerous and unknown-but-safe regions. Second, in the already-safe regions, cost gradients can harm reward prioritisation, stalling improvement. Lastly, under high $\lambda$ values, $Q_C$ saturates, marking nearly all actions as unsafe and stifling exploration.

These issues lead to a breakdown in the reward–cost trade-off: when $Q_C$ becomes uninformative under pressure, the Lagrangian mechanism degenerates, and learning stalls. This reveals a critical limitation and motivates the need for a safety critic that preserves gradient structure in cost. $Q_C$ must remain both conservative and sensitive enough to provide informative gradients, allowing the policy to balance reward and cost effectively. USC explicitly weights overestimation by the certainty of the critic, yielding sharper cost landscapes that support more scalable and precise trade-offs under the given constraints.


\section{USC Approach}
\label{sec:USC}

\begin{figure*}[tb]
    \centering
    \includegraphics[width=0.95\linewidth]{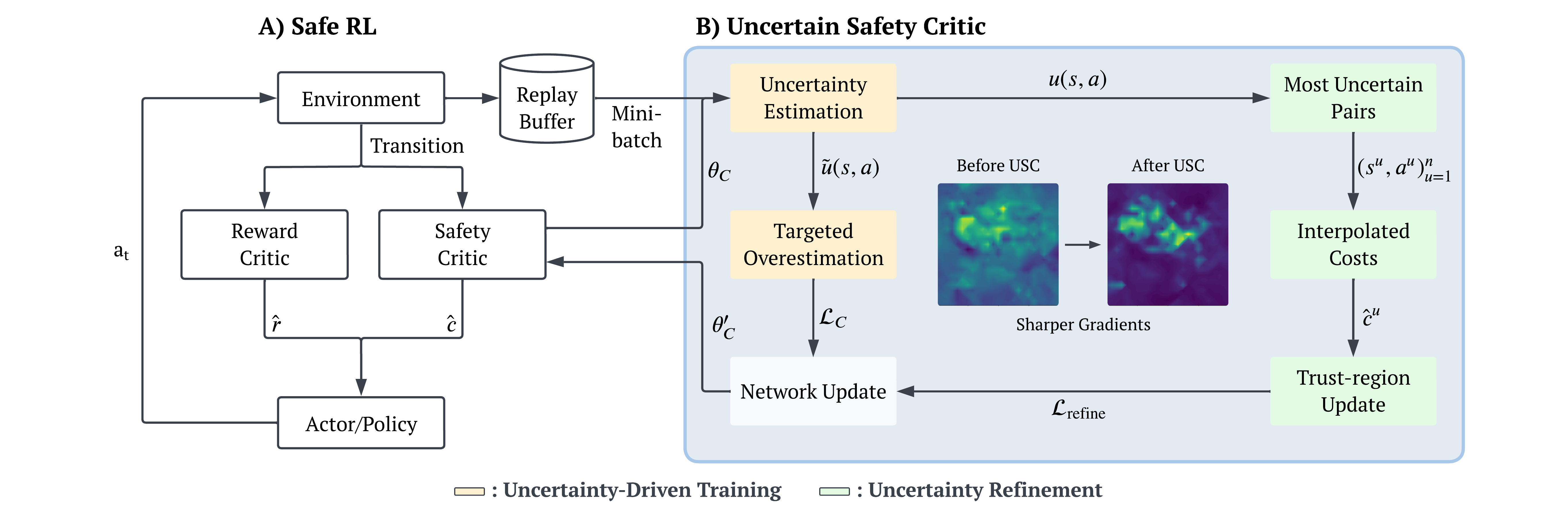}
    \vspace{-2mm}
    \caption{High-level USC overview, showing its integration into the safe RL training loop. The safety critic receives transitions from the environment and, alongside the reward critic, guides policy updates. Within USC, uncertainty estimation is used to compute influence scores $\tilde{u}(s,a)$, which drive targeted overestimation in the conservative loss $\mathcal{L}_C$ to preserve sharp, informative gradients. In parallel, the most uncertain state–action pairs are identified, interpolated with confident neighbours to form synthetic costs $\tilde{c}^u$, and incorporated into a trust-region update $\mathcal{L}_{\text{refine}}$. The Actor and Reward Critic Networks are updated as standard. Together, these mechanisms reduce over-conservatism while ensuring that gradient structure remains accurate and actionable. A detailed USC description is provided in Section~\ref{sec:USC}.}
    \vspace{-4mm}
    \label{fig:high-level-overview}
    \Description[cost map]{cost map}
\end{figure*}

To address the limitations of prior safety critics (Section~\ref{sec:Related Work}), USC introduces two key complementary mechanisms: 
an uncertainty-driven training objective that shapes the safety critic $Q_C$ using parameter-space uncertainty, and an uncertainty refinement procedure that actively reduces epistemic uncertainty in under-represented (low-data) regions of the state-action space. 
A high-level overview of USC is shown in Figure~\ref{fig:high-level-overview}.
During training, USC employs influence functions to estimate the sensitivity of critic parameters to specific training samples, quantifying uncertainty at the level of parameter attribution. 
These uncertainty estimates modulate the safety critic's cost predictions by reducing underestimation of costs in uncertain or costly regions and preventing their overestimation in adequately covered (safe) regions.
This distinction preserves conservative behaviour where needed, and maintains informative reward gradients from the reward critic $Q_R$.

To further improve the safety critic's reliability, USC identifies replay buffer samples (data) with the highest estimated uncertainty. 
For each such sample, USC performs linear interpolation with its nearest confidently predicted neighbour, generating synthetic samples that span the critic's decision boundary. 
These interpolated transitions are used in a constrained update to reduce uncertainty in sparsely represented regions.
We detail next both components, the uncertainty-aware training process and the refinement procedure, and explain the USC integration into the constrained RL loop.

\subsection{Uncertainty-Driven Training}
\label{sec:Training}

USC departs from widely-used failure-based safety formulations that rely on binary constraint signals $\{0,1\}$~\cite{bharadhwaj2020conservative,srinivasan2020learning}. Instead, USC operates under a continuous cost-based setting~\cite{achiam2017constrained}, where each transition incurs a real-valued cost signal $C(s,a) \in \mathbb{R}_+$. 
This formulation enables the safety critic to model both hard and soft constraint violations and is particularly well-suited to environments where constraint violations are graded, such as collision intensity, distance to hazard, or control effort. The safety critic predicts the expected cumulative cost under the current policy $\pi$. Unlike binary failure critics, the continuous cost formulation provides denser supervision, captures meaningful risk gradients essential for scalable safety under tighter constraints, and avoids ambiguity in defining failure, enhancing applicability across diverse safety settings.

The safety critic $Q_C$ estimates the cost of a particular action in a given state, and these estimates can be used to guide safer exploration. 
An ideal safety critic should yield accurate and conservative estimates as well as sharp and informative gradients. 
To create such a critic, we incorporate upper bound estimates on $Q_C$ to overestimate alternative actions from the policy and weight them based on
their confidence using Gauss-Newton influence~\cite{koh2017understanding}, thus preventing overestimation of well-predicted data.
We adopt Gauss-Newton over alternative influence functions because it provides stable, interpretable, and scalable uncertainty estimates using only first-order gradients and a positive semi-definite curvature~\cite{botev2017practical}. 

During updating the reward and safety critics,
for each data sample in the minibatch from the replay buffer $\mathcal{B} = \{(s_i, a_i)\}^B_{i=1}$, Gauss-Newton influence yields a scalar $u \geq 0$ that quantifies the safety critic's certainty about that data sample. The uncertainty scalar is given by 
\begin{equation}
\begin{aligned}
u(s,a) = \text{diag} &\Bigl( \bigl [\nabla_{\theta_C} Q_C(s, a; \theta^{*}_C) \bigr] \\
      &\times \Bigl(\sum_{i=1}^{B} \nabla_{\theta_C} Q_C(s_{i}, a_{i}; \theta^{*}_C) \nabla_{\theta_C} Q_C(s_{i}, a_{i}; \theta^{*}_C)^{\top} + \delta I \Bigr)^{-1} \\
      &\times \bigl[\nabla_{\theta_C} Q_C (s, a; \theta^{*}_C) \bigr]^{\top} \Bigr)
\label{eq:guass-newton-influence-safety-critic}
\end{aligned}
\end{equation}

where $\theta^*_C$ denotes the safety critic's $Q_C$ parameters (frozen weights) after the latest training update, and $\delta I$ is a small dampening term ensuring invertibility of the Gauss-Newton matrix.
Intuitively, $u$ measures the sensitivity of the safety critic's parameters are to each training sample; higher $u$ values indicate greater uncertainty, meaning that small changes to those samples would cause large shifts in the safety critic's parameters.
To integrate this uncertainty estimate into training, we modulate the standard conservative safety critic objective. Specifically, a Bellman regression objective is combined with a conservative penalty that incorporates the uncertainty $u$. For each transition $(s_t, a_t, r_t, s_{t+1}, c_t)$ sampled from the minibatch $\mathcal{B}$, we define the uncertainty-adjusted weight

\begin{equation}
\tilde{u}(s_t, a_t) = \log(1+ u(s_t, a_t))\times(1 + \mathbf{1}\{Q_C(s_t, a_t) > \frac{1}{|\mathcal{B}|}\sum^{|\mathcal{B}|}_{i=1}Q_C(s_i, a_i)\})
\label{eq:uncertainty-weight}
\end{equation}

The logarithmic transformation of $u$ ensures that influence values are stabilised, preventing extreme magnitudes from dominating the loss, while the indicator term accentuates penalties for transitions whose predicted cost exceeds the batch mean. 
Accordingly, conservatism is selectively concentrated on regions that are both uncertain and costly, rather than being spread uniformly across the state–action space. The safety critic is then updated by minimising
\begin{equation}
\begin{aligned}
&\mathcal{L}_C = \mathbb{E}_{(s_t, a_t, r_t, s_{t+1}, c_t)\sim\mathcal{B}}\Bigg[ 
    \frac{1}{2}\Big(Q_C(s_t, a_t) - \big(c_t + \gamma Q_C(s_{t+1}, \pi(s_{t+1}))\big)\Big)^2 \\
    &+ \frac{1}{2}\,\tilde{u}(s_t, a_t)\,
        \times \log 
            \sum_{a' \sim \text{Unif}(\mathcal{A})} 
            \exp\!\left( 
                \frac{Q_C(s_t, a') - \mu}{\sigma+\epsilon} 
            \right) - \frac{Q_C(s_t, a_t) - \mu}{\sigma+\epsilon}
\Bigg]
\label{eq:safety-critic-loss}
\end{aligned}
\end{equation}
where $\mu$ and $\sigma$ denote the batch mean and standard deviation of the safety critic outputs, and $\epsilon$ is a small constant for numerical stability. This objective improves $Q_C$ by shaping conservatism around epistemic uncertainty. The first term is a Bellman regression loss and drives $Q_C$ toward accuracy with respect to the observed cost signal $c_t$, ensuring consistency with the environment dynamics. 
The second term introduces an uncertainty-weighted conservative bias, yielding a higher loss when the log-sum-exp of $B$ randomly sampled alternatives appears safer than the chosen batch action. 
By amplifying this penalty in regions with high uncertainty or above-average costs, $\tilde{u}(s_t,a_t)$ ensures that the critic remains cautious 
where the consequences of underestimation are most severe. Conversely, it suppresses unnecessary inflation in well-sampled safe regions, preserving informative gradients that enable the policy to learn accurate reward–cost trade-offs. In this way, USC retains the robustness benefits of conservative safety critics while avoiding over-conservatism and creating sharp cost gradients to prevent stagnation in policy improvement.

\begin{theorem}
The log-exp-sum (LSE) penalty, used in Equation~\ref{eq:safety-critic-loss} and standard in CSC~\cite{bharadhwaj2020conservative}, is more effective at encouraging conservatism when the $m$ alternative actions are sampled uniformly over the action space, rather than from the current policy~\cite{bharadhwaj2020conservative}. If the policy aims to avoid high-cost regions, then the uniform sampling is more likely to draw from high-cost out-of-distribution actions. Formally, for any fixed state $s$, the conservative term is strictly larger under uniform scaling.
\label{thm:1}
\end{theorem}

The corresponding proof is in Appendix~\ref{sec:appendix-theo-analysis}. The safety critic's parameters are updated by
\begin{equation}
\theta_C \leftarrow \theta_C - \eta_C \nabla_{\theta_C} \mathcal{L}_C
\label{eq:safety-critic-params}
\end{equation}
where $\eta$ is the learning rate for the safety critic. In parallel, the reward critic $Q_R$ is trained and updated as standard~\cite{ddpg-paper}. The actor is optimised with a Lagrangian objective that balances reward maximisation, cost penalisation, and stability under policy shifts. Its loss is given by 
\begin{equation}
\begin{aligned}
\mathcal{L}_\pi 
&= -\,\mathbb{E}_{s_t \sim \mathcal{B}} \Big[ Q_R\big(s_t, \pi_{\theta_\pi}(s_t)\big) \Big] \\
&\quad + \lambda\,\mathbb{E}_{s_t \sim \mathcal{B}} \Big[ Q_C\big(s_t, \pi_{\theta_\pi}(s_t)\big) \Big] \\
&\quad + \delta\, D_{\mathrm{KL}}\!\Big(\pi_{\theta_\pi}(\cdot|s_t)\,\big\|\,\pi^{\text{old}}_{\theta_\pi}(\cdot|s_t)\Big)
\end{aligned}
\label{eq:actor-loss}
\end{equation}
and the parameters are updated by
\begin{equation}
\theta_\pi \leftarrow \theta_\pi - \eta_\pi \nabla_{\theta_\pi} \mathcal{L}_\pi
\label{eq:actor-params}
\end{equation}
The first term in the actor's loss $\mathcal{L}_\pi$ steers the policy toward actions with high expected reward, the second penalises actions with high predicted cost proportionally to a learnable cost coefficient $\lambda$, and the final KL term regularises deviations from the previous policy, ensuring stable optimisation even under strong safety pressure. 
After each episode, the dual variable $\lambda$ is updated based on the following equation to enforce the CMDP constraint 
%
\begin{equation}
\lambda \leftarrow \max\Big(0, \lambda - \eta_\lambda \big(\chi -\hat{C}\big)\Big)
\label{eq:dual-update}
\end{equation}
where $\hat{C}$ is the empirical average cost observed during training, $\eta_\lambda$ the learning rate, and $\chi$ the safety budget. This update tightens the penalty when constraints are violated and relaxes it otherwise, dynamically balancing reward pursuit and safety satisfaction.

All updates described above are embedded within the standard RL loop (Algorithm~\ref{alg:usc2} with the \fcolorbox{myblue}{myblue!20}{USC extension}). At each step, the actor proposes a candidate action, which is then screened by the safety critic. If the predicted cost exceeds that safety tolerance $\varepsilon_{\text{safe}}$, additional candidate actions are produced using Ornstein–Uhlenbeck noise~\cite{ddpg-paper}. Among these candidates, the action with the lowest predicted cost is selected, provided it satisfies the tolerance $\varepsilon_{\text{safe}}$; otherwise, the least risky action is chosen. This mechanism ensures that the policy not only maximises reward but also respects the safety critic’s guidance. 
Transitions are stored in the replay buffer, from which the actor and both critics are iteratively updated. The dual variable $\lambda$ is dynamically updated to maintain the cost budget. In this way, the safety critic is integral to both training and execution, shaping policy improvement while directly filtering unsafe actions.

{\centering
\begin{algorithm}[t]
\caption{Uncertain Safety Critic (USC) inside the Safe RL Loop}
\label{alg:usc2}
\begin{algorithmic}[1]
  \REQUIRE Environment $\mathcal{M}$, replay buffer $\mathcal{B}$, policy $\pi_{\theta_\pi}$, reward critic $Q_R$, safety critic $Q_C$, discount factor $\gamma$, safety budget $\chi$, dual variable $\lambda \geq 0$,  trust-region penalty coefficient $\beta$, trust threshold $\epsilon$, 
  learning rates $\eta_\pi,\eta_R,\eta_C,\eta_\lambda$, safety  tolerance $\epsilon_{\text{safe}}$
  \STATE Initialise $\theta_\pi,\theta_R,\theta_C$, targets, $\lambda \gets 1$
  \FOR{$e=1,2,\ldots,E_{\max}$}
    \STATE $s_1 \gets \textsc{Reset}()$, $t\gets 1$, \; episode cost accumulator $C_e \gets 0$
    \WHILE{$t \le T$ and $s_t$ not terminal}
      \STATE $a_t \gets \pi_{\theta_\pi}(s_t)$
      \begin{tcolorbox}[size=minimal, colback=myblue!10, colframe=myblue, boxrule=0.8pt, left=-2mm, right=-2mm, top=0pt, bottom=0pt, boxsep=5pt, before skip=0pt, after skip=0pt]
      \IF{$Q_C(s_t,a_t) > \epsilon_{\text{safe}}$}
        \STATE $\mathcal{N}(s_t) = \{a_t + \text{OU}(0, \sigma^2_{\text{OU}}), k=1,\ldots,N_{\text{samples}} \}$
        \STATE $a_t \gets \arg\min_{a' \in \{a_t\}\cup \mathcal{N}(s_t)} Q_C(s_t,a')$
      \ENDIF
      \end{tcolorbox}
      \STATE $(r_t,c_t,s_{t+1}) \gets \textsc{Step}(a_t)$; \quad $C_e \gets C_e + c_t$
      \STATE $\mathcal{B} \gets \mathcal{B} \cup \{(s_t,a_t,r_t,c_t,s_{t+1})\}$; \quad $t\gets t+1$
        \STATE $B \gets  \{(s_i,a_i,r_i,c_i,s'_i)\}_{i=1}^{|B|} \overset{\mathrm{iid}}{\sim} \mathcal{U}(\mathcal{B})$
        \begin{tcolorbox}[size=minimal, colback=myblue!10, colframe=myblue, boxrule=0.8pt, left=-2mm, right=-2mm, top=0pt, bottom=0pt, boxsep=5pt, before skip=0pt, after skip=0pt]
      \FOR{$\{(s_i,a_i,r_i,c_i,s'_i)\} \in \mathcal{B}$}
        \STATE $\!\!U \gets U \bigcup \textsc{CalculateUncertainty}(B, s_i, a_i)$ \hfill\COMMENT{Eq.~\ref{eq:guass-newton-influence-safety-critic}}
        \STATE $\!\!\tilde{U} \!\!\!\gets\! \tilde{U} \bigcup  \textsc{UncertaintyAdjustment}(B, u (s_i, a_i))$ \hfill\COMMENT{Eq.~\ref{eq:uncertainty-weight}}
      \ENDFOR
      
      \STATE $\mu \gets \frac{1}{B}\sum_{i} Q_C(s_i,a_i)$,\; \quad $\sigma \gets \mathrm{Std}\big(\{Q_C(s_i,a_i)\}_{i=1}^{|B|}, \mu\big)$

      \STATE $\mathcal{L}_C \gets \textsc{CalculateQ$_C$Loss}(B, \tilde{U}, \mu, \sigma)$  \hfill\COMMENT{Eq.~\ref{eq:safety-critic-loss}}
      \STATE $\theta_C \leftarrow \theta_C - \eta_C \nabla_{\theta_C} L_C$ \hfill\COMMENT{Eq.\ref{eq:safety-critic-params}}
       \end{tcolorbox}

      \STATE $\mathcal{L}_R \gets \textsc{CalculateQ$_R$Loss}(B, \gamma)$  
      \STATE $\theta_R \leftarrow \theta_R - \eta_R \nabla_{\theta_R} L_R$ 

      \STATE $L_\pi \gets \textsc{CalculateActorLoss} (B, \delta, \pi_{\theta_\pi} )  $ \hfill\COMMENT{Eq.~\ref{eq:actor-loss}}
      \STATE $\theta_\pi \leftarrow \theta_\pi - \eta_\pi \nabla_{\theta_\pi} L_\pi$ \hfill\COMMENT{Eq.\ref{eq:actor-params}}

      \begin{tcolorbox}[size=minimal, colback=myblue!10, colframe=myblue, boxrule=0.8pt, left=-2mm, right=-2mm, top=0pt, bottom=0pt, boxsep=5pt, before skip=0pt, after skip=0pt]
       $U^n \gets \textsc{SelectTop}(U, n)$
       \FOR{$u \in U^n$}
            \STATE $u^K \gets \textsc{FindNeighbours}(u, \mathcal{B}, K)  $
            \STATE $\hat{C}^u \gets \hat{C}^u \bigcup \textsc{CalculateTargetCost}(u^K)$ \hfill\COMMENT{Eq.\ref{eq:interp-cost}}
       \ENDFOR
       \STATE $\mathcal{L}_{\text{refine}} \gets \textsc{CalculateRefineLoss} (\hat{C}^u, U^n, \beta, \epsilon ) $ \hfil\COMMENT{Eq.~\ref{eq:refinement-loss}}

       \STATE $\theta_C \leftarrow \theta_C - \eta_C \nabla_{\theta_C} L_{\text{refine}}$ \hfill\COMMENT{Eq.\ref{eq:safety-critic-params-refine}}
      \end{tcolorbox}
    \ENDWHILE
    \STATE $\lambda \leftarrow \max\big(0,\; \lambda - \eta_\lambda\,(\chi - C_e)\big)$ \hfill\COMMENT{Eq.\ref{eq:dual-update}}
    \ENDFOR
\end{algorithmic}
\end{algorithm}
}

\subsection{Uncertainty Refinement}
\label{sec:Uncertainty Refinement}

While the uncertainty-weighted conservative objective in Section~\ref{sec:Training} encourages the safety critic to be more cautious 
in regions of increased epistemic uncertainty, it does not itself reduce that uncertainty. 
Specifically, poorly represented state–action regions can remain permanently over-penalised, leading to persistent over-conservatism and weak policy gradients. To address this, USC incorporates an uncertainty refinement step that actively targets the most uncertain samples and performs trust-region style updates to sharpen the safety critic’s predictions.

Concretely, after each gradient step, we identify a subset of the most uncertain samples in the replay buffer using the uncertainty scalar $u$ estimated via Gauss–Newton influence in Equation~\ref{eq:guass-newton-influence-safety-critic}. 
We rank the minibatch by $u(s_t, a_t)$ and select the top $n$ samples $U^n$. For each such sample, we generate synthetic cost targets via linear interpolation with nearby, confidently predicted replay samples. Specifically, let $(s^u, a^u)$ denote an uncertain state–action pair and $\{(s^k, a^k, c^k)\}_{k=1}^K$ its $K$ nearest neighbours in the joint state–action space. We define an interpolated target cost as
\begin{equation}
\begin{gathered}
\hat{c}^u = \sum^K_{k=1}w_k c^k \\
w_k = \frac{1/||(s^u, a^u) - (s^k, a^k)||}{\sum^K_{j=1}1/||(s^u, a^u) - (s^j, a^j)||}
\end{gathered}
\label{eq:interp-cost}
\end{equation}
where weights are inversely proportional to the distance in the state–action space. This interpolation encourages smoothness across the critic’s decision boundary, enabling uncertain regions to inherit structure from their nearest confident neighbours, thus reducing epistemic uncertainty in sparsely sampled regions.

To incorporate these synthetic targets, we adopt a trust region update on the safety critic's parameters. Given the old predictions $Q_C(s^u, a^u;\theta_C)$ and the new predictions under parameters $\theta_C'$, we compute a regularised loss
\begin{equation}
\begin{aligned}
\mathcal{L}_{\text{refine}} 
&= \frac{1}{|U^n|}\sum_{(s^u, a^u) \in U^n}\bigg[\frac{1}{2}\big(Q_C(s^u, a^u) - \hat{c}^u\big)^2 \\
&+ \beta \Big(\max\big\{0, D_{\text{KL}} \big(Q_C(s^u,a^u)|| Q_C(s^u,a^u;\theta_{\text{safe}}^{\text{old}})\big) - \epsilon \big\}\Big)^2 \bigg]
\end{aligned}
\label{eq:refinement-loss}
\end{equation}

\begin{table*}[tb]
\centering
\caption{Mean episodic reward and cost for comparative methods across evaluated benchmarks. \colorbox{lightblue}{Light blue} and \colorbox{lightgreen}{light green} highlighted cells indicate the lowest cost and highest reward for each task, respectively.}
\label{tab:main-results}
\resizebox{\textwidth}{!}{%
\begin{tabular}{@{}cccccccc@{}}
\toprule
 &  & \multicolumn{4}{c}{\textbf{Safety Gymnasium}} & \textbf{Gymnasium Robotics} & \textbf{Mujoco} \\ 
\cmidrule(lr){3-6} \cmidrule(lr){7-7} \cmidrule(lr){8-8} 
\textbf{Technique} & \textbf{Metric} & \textbf{Goal1 ($\chi = 1.0$)} & \textbf{Goal2 ($\chi = 5.0$)} & \textbf{Button1 ($\chi = 5.0$)} & \textbf{Button2 ($\chi = 5.0$)} & \textbf{FetchReach ($\chi = 5.0$)} & \textbf{HalfCheetah ($\chi = 0.1$)} \\ \midrule
\multirow{2}{*}{DDPG} & Reward $\uparrow$ & $6.57 \pm 0.47$ & $8.74 \pm 0.72$ & $6.20 \pm 0.74$ & $7.51 \pm 0.67$ & $-4.36 \pm 16.04$ & $5.71 \pm 2.48$ \\
 & Cost $\downarrow$ & $1.05 \pm 0.39$ & $5.36 \pm 1.47$ & $9.50 \pm 3.01$ & $10.30 \pm 3.32$ & $8.19 \pm 7.70$ & $21.82 \pm 23.20$ \\ 
\multirow{2}{*}{PID-Lag} & Reward $\uparrow$ & $6.48 \pm 0.49$ & $5.88 \pm 3.39$ & $2.49 \pm 2.57$ & $1.30 \pm 2.61$ & $-1.43 \pm 5.96$ & $4.69 \pm 1.52$ \\
 & Cost $\downarrow$ & $1.14 \pm 0.51$ & \cellcolor{lightblue}\bm{$4.94 \pm 2.27$} & \cellcolor{lightblue}\bm{$4.50 \pm 4.26$} & \cellcolor{lightblue}\bm{$3.40 \pm 4.77$} & $3.45 \pm 3.12$ & $0.36 \pm 0.35$ \\
\multirow{2}{*}{CPO} & Reward $\uparrow$ & $6.40 \pm 0.76$ & $8.74 \pm 0.81$ & $5.99 \pm 1.20$ & $7.56 \pm 0.79$ & $1.80 \pm 4.49$ & $3.69 \pm 1.89$ \\
 & Cost $\downarrow$ & $1.21 \pm 0.51$ & $5.08 \pm 1.26$ & $7.47 \pm 2.62$ & $9.05 \pm 2.80$ & $4.24 \pm 1.70$ & $1.71 \pm 1.93$ \\ 
\multirow{2}{*}{RCPO} & Reward $\uparrow$ & $6.56 \pm 0.38$ & $8.28 \pm 0.95$ & $4.21 \pm 2.38$ & $3.67 \pm 3.18$ & $-0.77 \pm 6.57$ & $6.28 \pm 2.13$ \\
 & Cost $\downarrow$ & $1.02 \pm 0.40$ & $5.13 \pm 1.71$ & $5.46 \pm 3.35$ & $4.58 \pm 3.72$ & $3.84 \pm 2.36$ & $0.94 \pm 0.84$ \\ 
\multirow{2}{*}{SC} & Reward $\uparrow$ & \cellcolor{lightgreen}\bm{$6.60 \pm 0.45$} & $7.54 \pm 1.38$ & $5.12 \pm 1.89$ & $2.48 \pm 3.32$ & $1.98 \pm 4.65$ & $3.19 \pm 3.43$ \\
 & Cost $\downarrow$ & $1.02 \pm 0.37$ & $5.65 \pm 2.00$ & $5.96 \pm 2.68$ & $4.04 \pm 4.46$ & \cellcolor{lightblue}\bm{$3.18 \pm 1.05$} & $0.32 \pm 0.63$ \\ 
\multirow{2}{*}{CSC} & Reward $\uparrow$ & $6.56 \pm 0.40$ & $8.65 \pm 0.75$ & $5.48 \pm 1.83$ & $6.45 \pm 2.44$ & $-1.53 \pm 6.51$ & $4.13 \pm 1.29$ \\
 & Cost $\downarrow$ & $1.03 \pm 0.39$ & $5.28 \pm 1.44$ & $5.57 \pm 2.63$ & $6.49 \pm 3.19$ & $3.91 \pm 3.07$ & $2.71 \pm 5.70$ \\ 
\multirow{2}{*}{USC (NR)} & Reward $\uparrow$ & $6.56 \pm 0.39$ & $8.74 \pm 0.78$ & $6.31 \pm 0.55$ & $7.49 \pm 0.79$ & $0.55 \pm 4.00$ & \cellcolor{lightgreen}\bm{$6.64 \pm 2.26$} \\
 & Cost $\downarrow$ & $1.06 \pm 0.41$ & $5.53 \pm 1.43$ & $6.37 \pm 2.05$ & $8.02 \pm 2.52$ & $3.34 \pm 1.36$ & $0.62 \pm 0.51$ \\ 
\multirow{2}{*}{USC} & Reward $\uparrow$ & $6.55 \pm 0.39$ & \cellcolor{lightgreen}\bm{$8.84 \pm 0.62$} & \cellcolor{lightgreen}\bm{$6.38 \pm 0.62$} & \cellcolor{lightgreen}\bm{$7.69 \pm 0.62$} & \cellcolor{lightgreen}\bm{$3.92 \pm 2.51$} & $3.76 \pm 2.70$ \\
 & Cost $\downarrow$ & \cellcolor{lightblue}\bm{$0.96 \pm 0.33$} & $5.05 \pm 1.27$ & $6.02 \pm 2.13$ & $6.32 \pm 2.23$ & $3.20 \pm 1.11$ & \cellcolor{lightblue}\bm{$0.20 \pm 0.40$} \\ 
 \bottomrule
\end{tabular}
}
\end{table*}

where the first term enforces consistency with interpolated costs, and the second penalises large deviations from the previous critic predictions beyond a threshold $\epsilon$, and $\beta$ is a trust-region penalty coefficient controlling the strength of the KL regularisation term. The quadratic penalty ensures that updates remain local in function space, preventing destabilising shifts in low-data regions. 

Since refinement is performed using synthetic, interpolated costs rather than direct environment feedback, we adopt a trust-region update to guard against instability. This ensures that, even if interpolated values occasionally misalign with the true dynamics, the resulting updates remain local and cannot distort the safety critic. The critic parameters are then refined by
\begin{equation}
\theta_C \leftarrow \theta_C - \eta_C \nabla_{\theta_C} \mathcal{L}_{\text{refine}} 
\label{eq:safety-critic-params-refine}
\end{equation}
This refinement has two benefits. First, by interpolating uncertain points with nearby confident samples, it reduces epistemic uncertainty in sparse regions, preventing the agent from avoiding areas that may contain local or global optima. Second, the trust-region penalty stabilises learning by preventing abrupt shifts in cost estimates, ensuring that conservatism does not collapse into overconfidence. These mechanisms jointly enable USC to sharpen cost predictions where uncertainty is most significant, mitigating the known over-conservatism of CSC while maintaining a principled view toward safety.

\begin{figure}[tb!]
    \centering
    \includegraphics[width=\linewidth]{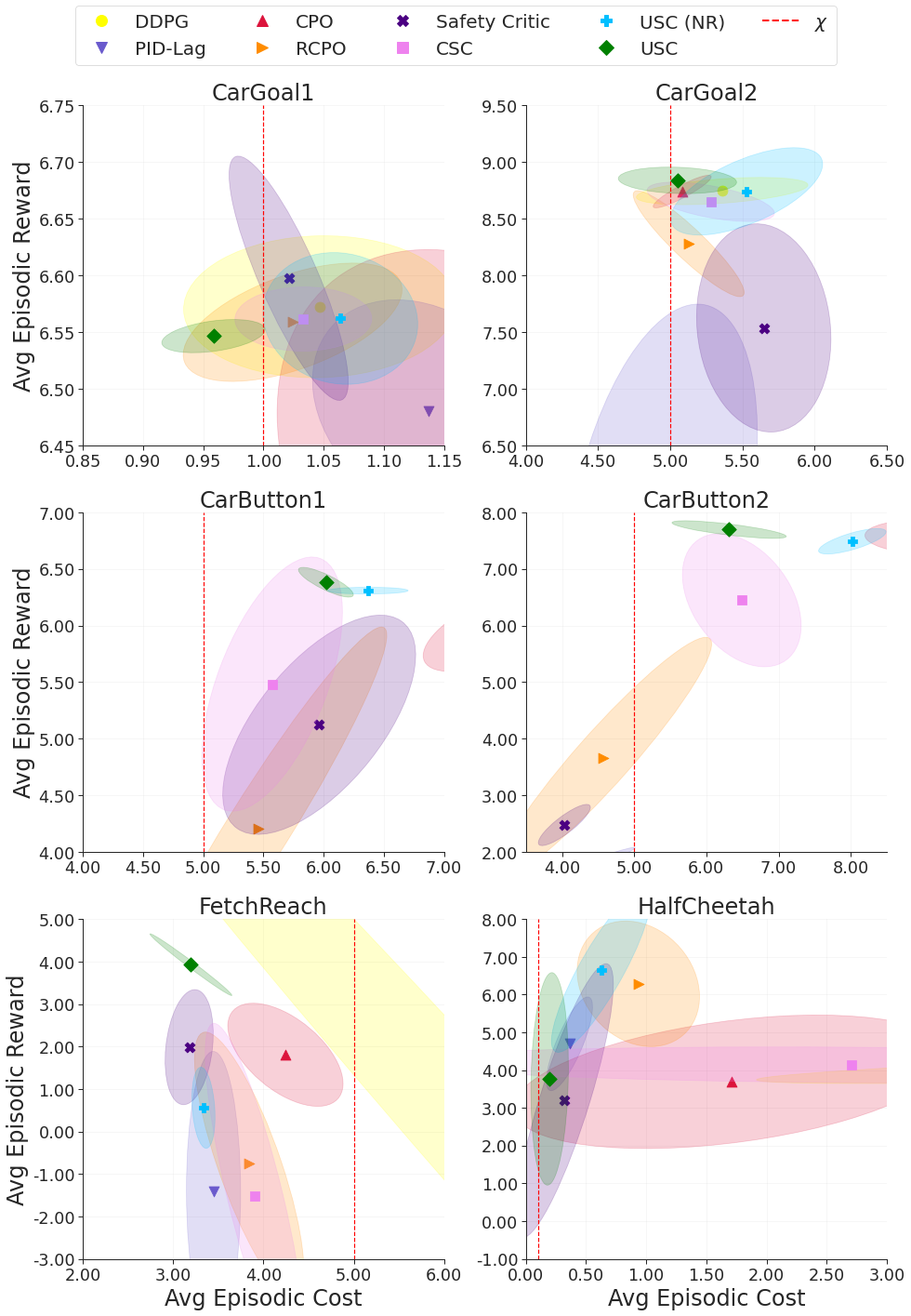}
    \vspace{-6mm}
    \caption{Pareto plot of average episodic cost versus average episodic reward across evaluated benchmarks. Each point denotes the mean performance of a method, with shaded ellipses representing standard deviation.}
    \label{fig:main-results-pareto}
    \Description[Pareto plot showing reward vs. cost trade-offs, with points as mean performance and ellipses as variance.]{Pareto plot showing reward vs. cost trade-offs, with points as mean performance and ellipses as variance.}
    \vspace{-4mm}
\end{figure}


\section{Evaluation}
\label{sec:Evaluation}

We evaluate USC~\footnote{Code available at: https://github.com/team-daniel/USC} using tasks from the Safety Gymnasium~\cite{ji2023safety}, Gymnasium Robotics~\cite{gymrobot}, and Mujoco~\cite{towers2024gymnasium} test-suites. These assess performance and safety in high-dimensional scenarios with various constraints, enabling a fair comparison of USC against approaches that also use similar benchmarks~\cite{bharadhwaj2020conservative}. The selected tasks cover varying complexities in constraints, goals, and agent types:

\begin{itemize}[noitemsep, nolistsep, leftmargin=3mm]
    \item \textbf{CarGoal1}: A car-like wheeled agent must reach a goal while avoiding 9 hazards, all with randomised positions.
    \item \textbf{CarGoal2}: Similar to CarGoal1, but with 20 hazards.
    \item \textbf{CarButton1}: An agent must press the correct button, avoiding 8 hazards and 3 incorrect buttons, all with randomised positions.
    \item \textbf{CarButton2}: Similar to the CarButton1, with 14 hazards.
    \item \textbf{FetchReach}: An agent controls a 7-DoF robot’s end effector to move to a random target position in 3D space, both with randomised positions.
    \item \textbf{HalfCheetah}: A bipedal agent must run forward while maintaining a safe velocity.
\end{itemize}

Details of these environments and tasks are in Appendix~\ref{sec:appendix-task-details}; Figure~\ref{fig:appendix-task-vis} shows task instances for visualisation.

We perform a detailed evaluation of USC and compare it with the following state-of-the-art RL algorithms:
\begin{itemize}[noitemsep, nolistsep, leftmargin=3mm]	
    \item \textbf{DDPG}: A deep deterministic policy gradient agent ~\cite{ddpg-paper} which is  the foundational baseline.
    \item \textbf{PID-Lag}: A Lagrangian constraint penalty using PID control to stabilise training~\cite{stooke2020responsive}.
    \item \textbf{CPO}: A trust-region method that enforces constraints during each policy update~\cite{achiam2017constrained}.
    \item \textbf{RCPO}: A method which incorporates constraint costs into the reward via a learned Lagrange multiplier~\cite{tessler2018reward}.
    \item \textbf{Safety Critic (CSC)}: A DDPG agent leveraging a safety critic to evaluate safety~\cite{srinivasan2020learning}. 
    \item \textbf{Conservative Safety Critic (CSC)}: A DDPG agent with a conservative safety critic that uses conservative estimates to evaluate safety~\cite{bharadhwaj2020conservative}. 
\end{itemize}

We also ablate USC by comparing against Uncertain Safety Critic (NR), a USC variant without the uncertainty refinement stage. Other safe exploration approaches discussed in Section~\ref{sec:Related Work} were excluded because they represent earlier research from approaches used in our evaluation. To enable a fair comparison, 
all methods use the same DDPG configuration; algorithm details and hyperparameter information are provided in Appendix~\ref{sec:appendix-hyperparams}. Following~\cite{bharadhwaj2020conservative}, we perform five independent runs, with results showing mean scores and confidence intervals (standard error).

\subsection{Performance Results}
\label{sec:core-results}
Table~\ref{tab:main-results} shows the average episodic reward and cost for all comparative methods in the evaluated benchmarks. Figure~\ref{fig:main-results-pareto} additionally highlights the cost-reward trade-off seen during training. Full results including training plots are in Appendix~\ref{sec:appendix-full-results}. The results show that USC effectively balances reward and cost better than any other method across all tasks.

The primary role of a safety critic is to provide informative cost signals that enable policies to balance safety and performance. Across all benchmarks, the base agent (DDPG) cannot ensure safety at all, since it lacks any safety mechanism and optimizes purely for reward, leading to consistently high violation rates. USC consistently reduces safety violations while supporting stronger reward learning. For example, in CarGoal2, USC lowers the average episodic cost from $5.65 \pm 2.00$ with a standard safety critic to $5.05 \pm 1.27$, while simultaneously achieving higher rewards $8.84 \pm 0.62$ compared to both SC $7.54  \pm 1.38$ and CSC $8.65 \pm 0.75$. 
In CarButton1, USC reduces cost by nearly $40\%$ relative to DDPG, achieving the highest overall reward $6.38 \pm 0.62$ and a very competitive cost compared to the best result achieved by CSC. 
In larger and more complex environments such as FetchReach, USC’s scalability is most evident. In this environment, where CSC yielded a high cost $3.91 \pm 3.07$  and a poor reward of $-1.53 \pm 6.51$, USC achieves the highest reward while maintaining nearly the lowest cost, matching the standard SC’s cost level but delivering roughly double the reward.

\begin{table}[tb]
\centering
\caption{Predictive cost map errors in CarGoal2 compared to ground truth, showing gradient alignment, safe–unsafe contrast, and hazard boundary sharpness.}
\vspace{-4mm}
\label{tab:costmap-metrics}
\resizebox{\linewidth}{!}{%
\begin{tabular}{@{}lccc@{}}
\toprule
 & \multicolumn{3}{c}{Method} \\ \cmidrule(l){2-4}
Metric & Safety Critic & Conservative Safety Critic & Uncertain Safety Critic \\ 
\midrule
Gradient MSE $\downarrow$ & $0.58 \pm 0.29$ & $1.95 \pm 1.37$ & \cellcolor{lightblue}\bm{$0.10 \pm 0.06$} \\
Contrast Error $\downarrow$ & $0.09 \pm 0.09$ & \cellcolor{lightblue}\bm{$0.04 \pm 0.03$} & $0.08 \pm 0.05$ \\
Entropy Error $\downarrow$ & $2.50 \pm 0.67$ & $3.02 \pm 0.27$ & \cellcolor{lightblue}\bm{$2.16 \pm 0.26$} \\ 
\bottomrule
\vspace{-8mm}
\end{tabular}%
}
\end{table}

\begin{figure}[tb!]
    \centering
    \includegraphics[width=\linewidth]{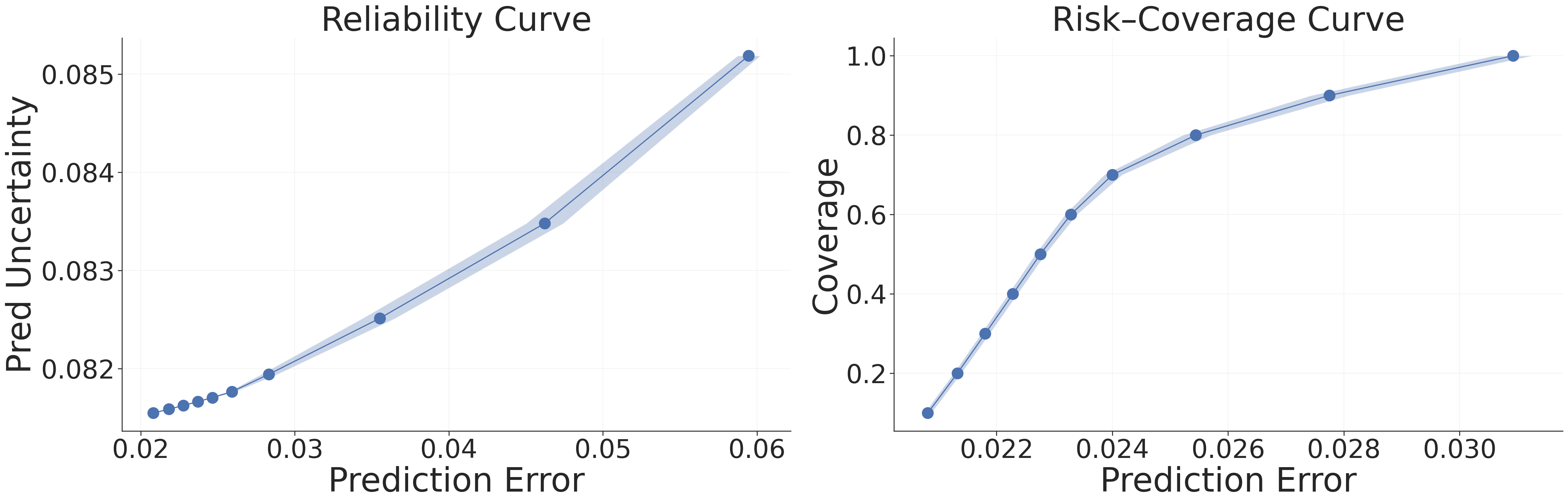}
    \vspace{-4mm}
    \caption{Reliability curve (left) showing prediction error vs predicted uncertainty, and risk–coverage curve (right) showing prediction error as a function of coverage ordered by uncertainty, where lower coverage corresponds to keeping only the most confident predictions.}
    \label{fig:reliability-risk-coverage-plot}
    \Description[Reliability and risk–coverage curves linking prediction error to uncertainty and coverage for USC.]{Reliability and risk–coverage curves linking prediction error to uncertainty and coverage for USC.}
    \vspace{-4mm}
\end{figure}

\begin{table*}[tb]
\centering
\caption{Mean episodic reward and episodic cost for comparative methods across evaluated benchmarks during 10 independent deployment runs. \colorbox{lightblue}{Light blue} and \colorbox{lightgreen}{light green} highlighted cells indicate the lowest cost and highest reward for each task, respectively. \colorbox{lightred}{Light red} highlighted cells denote models that do not achieve 100\% success rate.}
\label{tab:usc-deployment-results}
\resizebox{\textwidth}{!}{%
\begin{tabular}{@{}cccccccc@{}}
\toprule
 &  & \multicolumn{4}{c}{\textbf{Safety Gymnasium}} & \textbf{Gymnasium Robotics} & \textbf{Mujoco} \\ 
\cmidrule(lr){3-6} \cmidrule(lr){7-7} \cmidrule(lr){8-8} 
\textbf{Technique} & \textbf{Metric} & \textbf{Goal1 ($\chi = 1.0$)} & \textbf{Goal2 ($\chi = 5.0$)} & \textbf{Button1 ($\chi = 5.0$)} & \textbf{Button2 ($\chi = 5.0$)} & \textbf{FetchReach ($\chi = 5.0$)} & \textbf{HalfCheetah ($\chi = 0.1$)} \\ \midrule
\multirow{2}{*}{DDPG} & Reward $\uparrow$ & \cellcolor{lightgreen}\bm{$7.90 \pm 2.40$} & $9.13 \pm 2.88$ & \cellcolor{lightred}\bm{$5.36 \pm 1.87$} & $9.09 \pm 2.69$ & $4.74 \pm 0.10$ & $9.24 \pm 2.17$ \\
 & Cost $\downarrow$ & $3.10 \pm 4.11$ & $41.80 \pm 40.59$ & \cellcolor{lightred}\bm{$71.40 \pm 126.01$} & $31.30 \pm 24.58$ & $3.09 \pm 1.22$ &  $69.80 \pm 27.32$\\ 
\multirow{2}{*}{PID-Lag} & Reward $\uparrow$ & $6.50 \pm 2.11$ & $8.73 \pm 3.51$ & \cellcolor{lightred}\bm{$0.07 \pm 0.19$} & \cellcolor{lightred}\bm{$0.07 \pm 0.20$} & \cellcolor{lightred}\bm{$3.68 \pm 3.06$} & $5.21 \pm 0.47$ \\
 & Cost $\downarrow$ & $4.60 \pm 7.00$ & $25.10 \pm 16.59$ & \cellcolor{lightred}\bm{$0.00 \pm 0.00$} & \cellcolor{lightred}\bm{$0.00 \pm 0.00$} & \cellcolor{lightred}\bm{$2.61 \pm 1.73$} & \cellcolor{lightblue}\bm{$0.20 \pm 0.40$} \\
\multirow{2}{*}{CPO} & Reward $\uparrow$ & $6.76 \pm 2.52$ & $8.04 \pm 2.98$ & $6.22 \pm 2.06$ & $7.73 \pm 4.52$ & $4.80 \pm 0.11$ & $6.57 \pm 0.26$ \\
 & Cost $\downarrow$ & $7.80 \pm 11.74$ & $21.50 \pm 20.41$ & $43.70 \pm 64.24$ & $47.30 \pm 84.94$ & $2.80 \pm 1.17$ & $10.90 \pm 3.59$ \\ 
\multirow{2}{*}{RCPO} & Reward $\uparrow$ & $6.74 \pm 2.21$ & $8.02 \pm 3.76$ & \cellcolor{lightred}\bm{$-0.01 \pm 0.26$} & \cellcolor{lightred}\bm{$-5.30 \pm 12.53$} & $4.72 \pm 0.16$ & \cellcolor{lightgreen}\bm{$9.48 \pm 1.68$} \\
 & Cost $\downarrow$ & $2.40 \pm 4.98$ & $33.20 \pm 53.33$ & \cellcolor{lightred}\bm{$0.00 \pm 0.00$} & \cellcolor{lightred}\bm{$101.80 \pm 147.89$} & $2.53 \pm 1.61$ & $0.80 \pm 1.17$ \\ 
\multirow{2}{*}{SC} & Reward $\uparrow$ & $7.57 \pm 1.89$ & \cellcolor{lightgreen}\bm{$11.48 \pm 2.51$} & \cellcolor{lightred}\bm{$2.63 \pm 2.14$} & \cellcolor{lightred}\bm{$0.11 \pm 0.17$} & \cellcolor{lightgreen}\bm{$4.81 \pm 0.09$} & $6.91 \pm 0.47$ \\
 & Cost $\downarrow$ & $4.80 \pm 5.93$ & $31.20 \pm 33.28$ & \cellcolor{lightred}\bm{$0.00 \pm 0.00$} & \cellcolor{lightred}\bm{$0.00 \pm 0.00$} & $3.12 \pm 1.09$ & $0.50 \pm 0.92$ \\ 
\multirow{2}{*}{CSC} & Reward $\uparrow$ & $7.21 \pm 2.64$ & $9.44 \pm 2.79$ & \cellcolor{lightred}\bm{$3.96 \pm 1.62$} & \cellcolor{lightred}\bm{$6.33 \pm 2.95$} & $4.78 \pm 0.10$ & $8.66 \pm 1.57$ \\
 & Cost $\downarrow$ & $2.00 \pm 6.00$ & $27.00 \pm 17.49$ & \cellcolor{lightred}\bm{$7.90 \pm 15.93$} & \cellcolor{lightred}\bm{$14.10 \pm 17.39$} & $2.98 \pm 1.17$ & $16.00 \pm 10.11$ \\ 
\multirow{2}{*}{USC (NR)} & Reward $\uparrow$ & $5.79 \pm 2.34$ & $9.38 \pm 3.95$ & \cellcolor{lightgreen}\bm{$7.65 \pm 2.66$} & \cellcolor{lightgreen}\bm{$9.59 \pm 2.59$} & $4.78 \pm 0.13$ & $5.74 \pm 0.28$ \\
 & Cost $\downarrow$ & $2.60 \pm 5.66$ & $26.30 \pm 26.71$ & $22.20 \pm 25.96$ & $24.00 \pm 41.23$ & $2.97 \pm 1.75$ & $0.60 \pm 0.49$ \\ 
\multirow{2}{*}{USC} & Reward $\uparrow$ & $7.19 \pm 1.97$ & $9.48 \pm 3.30$ & $6.41 \pm 2.53$ & $ 8.51 \pm 4.79$ & $4.78 \pm 0.07$ & $5.44 \pm 0.42$ \\
 & Cost $\downarrow$ & \cellcolor{lightblue}\bm{$0.70 \pm 2.10$} & \cellcolor{lightblue}\bm{$20.30 \pm 24.63$} & \cellcolor{lightblue}\bm{$15.90 \pm 28.40$} & \cellcolor{lightblue}\bm{$13.60 \pm 13.71$} & \cellcolor{lightblue}\bm{$2.45 \pm 1.25$} & \cellcolor{lightblue}\bm{$0.20 \pm 0.40$} \\ 
 \bottomrule
\end{tabular}
}
\end{table*}

\begin{table}[tb]
\centering
\caption{Success rates (\%) for comparative methods across evaluated benchmarks in 10 independent deployment runs.\colorbox{lightred}{Light red} highlighted cells denote models that do not achieve 100\% success rate.}
\vspace{-4mm}
\label{tab:test-time-results-success}
\resizebox{\linewidth}{!}{%
\begin{tabular}{@{}ccccccc}
\toprule
\cmidrule(lr){2-5} \cmidrule(lr){6-6} \cmidrule(lr){7-7}
\textbf{Technique} & \textbf{Goal1} & \textbf{Goal2} & \textbf{Button1} & \textbf{Button2} & \textbf{FetchReach} \\ \midrule
DDPG & $100\%$ & $100\%$ & \cellcolor{lightred}\bm{$90\%$} & $100\%$ & $100\%$ \\ 
PID-Lag & $100\%$ & $100\%$ & \cellcolor{lightred}\bm{$0\%$} & \cellcolor{lightred}\bm{$0\%$} & \cellcolor{lightred}\bm{$90\%$} \\ 
CPO & $100\%$ & $100\%$ & $100\%$ & $100\%$ & $100\%$ \\ 
RCPO & $100\%$ & $100\%$ & \cellcolor{lightred}\bm{$0\%$} & \cellcolor{lightred}\bm{$30\%$} & $100\%$ \\ 
SC & $100\%$ & $100\%$ & \cellcolor{lightred}\bm{$50\%$} & \cellcolor{lightred}\bm{$0\%$} & $100\%$ \\ 
CSC & $100\%$ & $100\%$ & \cellcolor{lightred}\bm{$70\%$} & \cellcolor{lightred}\bm{$80\%$} & $100\%$ \\ 
USC (NR) & $100\%$ & $100\%$ & $100\%$ & $100\%$ & $100\%$ \\ 
USC & $100\%$ & $100\%$ & $100\%$ & $100\%$ & $100\%$ \\ 
\bottomrule
\end{tabular}
}
\end{table}

To further disentangle the role of USC’s refinement step, we evaluate an ablation variant, USC (NR), which omits uncertainty refinement and relies solely on uncertainty-weighted overestimation. Without refinement, USC (NR) is still able to achieve reward levels comparable to the full USC, since the uncertainty-modulated conservative loss preserves informative gradients that guide effective reward optimisation. However, the absence of refinement leaves highly uncertain regions poorly constrained, leading to inflated or diffuse cost estimates in sparsely sampled parts of the state–action space. Consequently, USC (NR) struggles to consistently keep costs within budget, highlighting that the refinement step is essential for stabilising safety by reducing epistemic uncertainty where the critic lacks reliable coverage.

These results are further supported by predictive cost map error metrics shown in Table~\ref{tab:costmap-metrics}. Gradient MSE quantifies alignment between predicted cost gradients and the true hazard boundaries. USC achieves a score of $0.10 \pm 0.06$, an improvement of over $80\%$ compared to SC ($0.58 \pm 0.29$) and an order of magnitude larger than CSC ($1.95 \pm 1.37$), confirming that USC preserves meaningful gradient structure rather than collapsing into flat or saturated predictions. Contrast Error measures the critic’s ability to separate hazardous from safe regions. While CSC achieves the lowest value with $0.04 \pm 0.03$, it does so by uniformly inflating cost across the state–action space, whereas USC achieves nearly comparable separation ($0.08 \pm 0.05$) without sacrificing gradient fidelity, thus maintaining useful trade-off signals. Finally, Entropy Error captures how concentrated predicted hazards are relative to ground truth. USC achieves the lowest value of $2.16 \pm 0.26$ among all methods, indicating sharper and less diffuse hazard boundaries. Together, these metrics highlight that USC produces cost maps that are simultaneously sharper, better aligned with hazards, and more informative for guiding safe policy optimisation.

Figure~\ref{fig:reliability-risk-coverage-plot} shows the reliability and risk–coverage curves for USC’s uncertainty estimates. The reliability curve (top) plots prediction error against predicted uncertainty. Its near-monotonic trend indicates that higher predicted uncertainty corresponds to higher error, demonstrating that the critic’s uncertainty estimates are well-calibrated and meaningfully reflect the reliability of its cost predictions. Additionally, the risk–coverage curve (bottom) shows how prediction error varies as we restrict evaluation to subsets of predictions with the highest confidence. As coverage decreases—focusing only on the most certain predictions, the error consistently drops, confirming that uncertainty values can be used to selectively filter reliable estimates. Together, these curves highlight that USC’s uncertainty estimates are not only calibrated but also actionable, providing a principled basis for modulating conservatism and refining predictions in regions of high epistemic uncertainty.

\subsection{Deployment Results}
\label{sec:test-time-results}

Table~\ref{tab:usc-deployment-results} shows deployment performance (mean episodic reward and episodic cost) across the benchmark tasks, with success rates shown in Table~\ref{tab:test-time-results-success}. The base agent (DDPG) achieves relatively high rewards but at the expense of consistently high costs, confirming that without any safety mechanism, it optimises purely for performance and cannot ensure constraint satisfaction. Standard safety critics reduce costs to some extent but remain unstable, exhibiting frequent violations due to unreliable gradient structure, often failing to complete more complex tasks reliably. CSC exhibits the opposite tendency, i.e., although it lowers costs more aggressively, its excessive conservatism substantially degrades reward, thus failing to solve certain environments (e.g., CarButtong1, CarButton2).

In contrast, USC maintains a stronger balance. Across multiple environments, it matches or exceeds the reward levels of DDPG while simultaneously reducing costs closer to budget limits. For example, in CarGoal2, USC achieves both near-optimal reward and competitive costs compared to SC, demonstrating its capability to  enforce constraints without collapsing performance. Similarly, in CarButton1, USC achieves the highest reward among all methods while also achieving one of the lowest costs, highlighting its ability to leverage uncertainty-aware modulation to encourage efficient yet safe exploration. Importantly, unlike SC and CSC, USC consistently succeeds in solving complex tasks, achieving a 100\% success rate, clearly reflecting its scalability and robustness under deployment.

Taken together, these results provide concrete evidence that USC scales effectively to deployment settings, providing a principled trade-off between reward, safety, and task completion that neither standard nor conservative critics can achieve.


\section{Conclusion}
\label{sec:Conclusion}

USC is an uncertainty-aware safety critic that mitigates the over-conservatism of existing safety critics by enforcing a sharp and accurate gradient structure through uncertainty-aware modulation and refinement. This unique USC feature enables the agent's policy to maintain an effective balance between cost and reward, even as task complexity and scale increase. Our evaluation demonstrates that USC consistently reduces safety violations while maintaining or exceeding state-of-the-art performance, particularly in larger, complex environments. Future work will investigate extending USC to support partially observable and adversarial environments.

\printbibliography


\newpage
\onecolumn
\appendix
\begingroup
\raggedbottom


\section{Task Details}
\label{sec:appendix-task-details}
We evaluate USC and the comparative methods in various tasks across three separate test suites. In our experiments, found in Section~\ref{sec:Evaluation}, we chose to use:

\begin{figure}[htb]
\centering
\begin{minipage}{0.65\textwidth}
    \begin{minipage}[t]{0.24\linewidth}
        \centering
        \includegraphics[width=\linewidth]{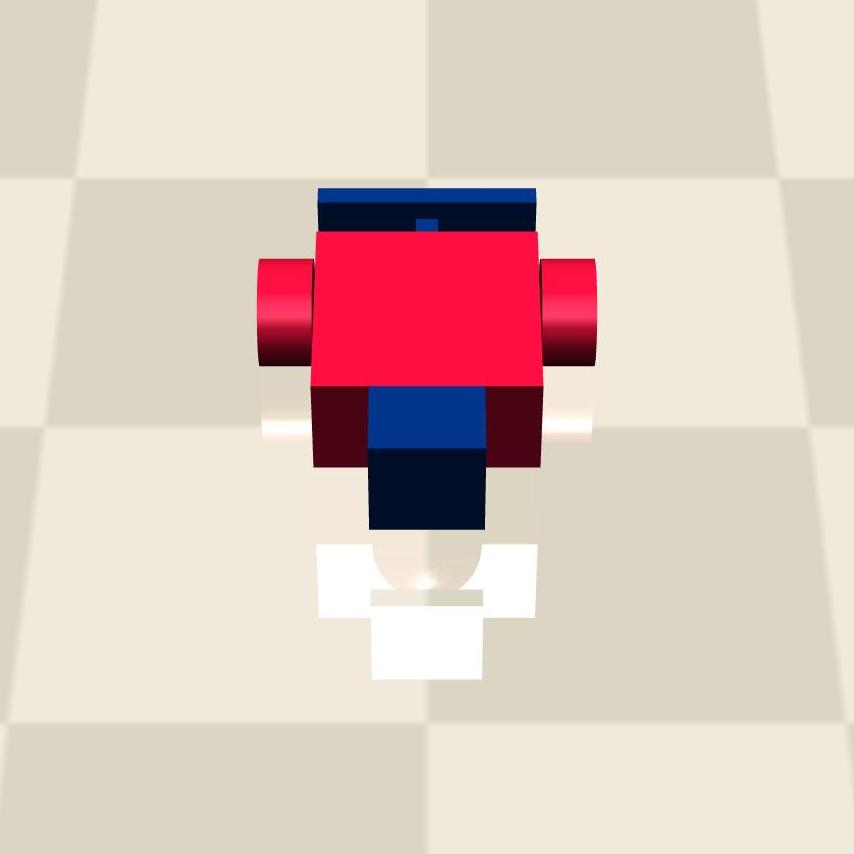}
        \smallskip
        \textbf{(a)} Car
    \end{minipage}
    \hfill
    \begin{minipage}[t]{0.24\linewidth}
        \centering
        \includegraphics[width=\linewidth]{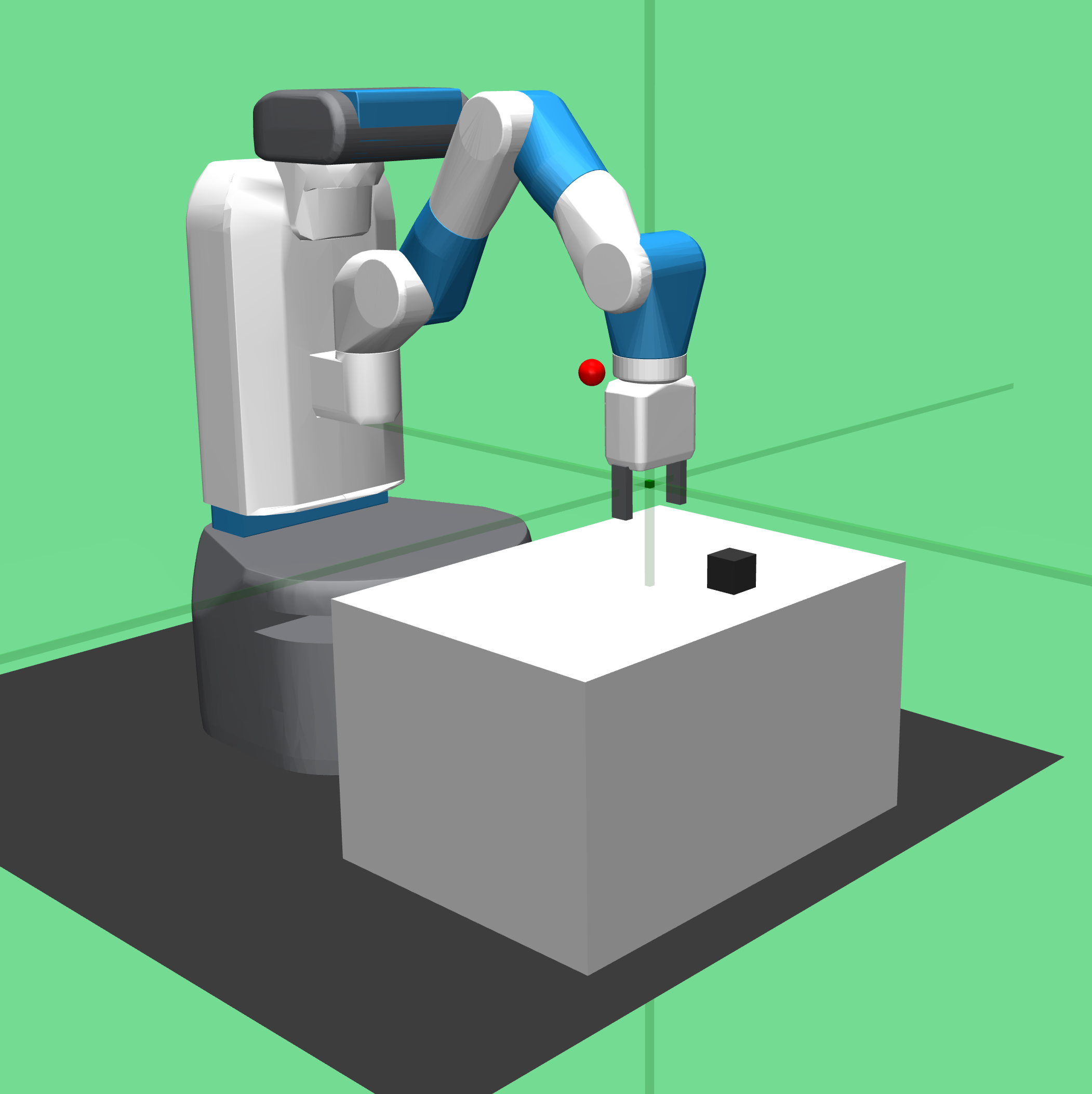}
        \smallskip
        \textbf{(b)} Fetch Manipulator
    \end{minipage}
    \hfill
    \begin{minipage}[t]{0.24\linewidth}
        \centering
        \includegraphics[width=\linewidth]{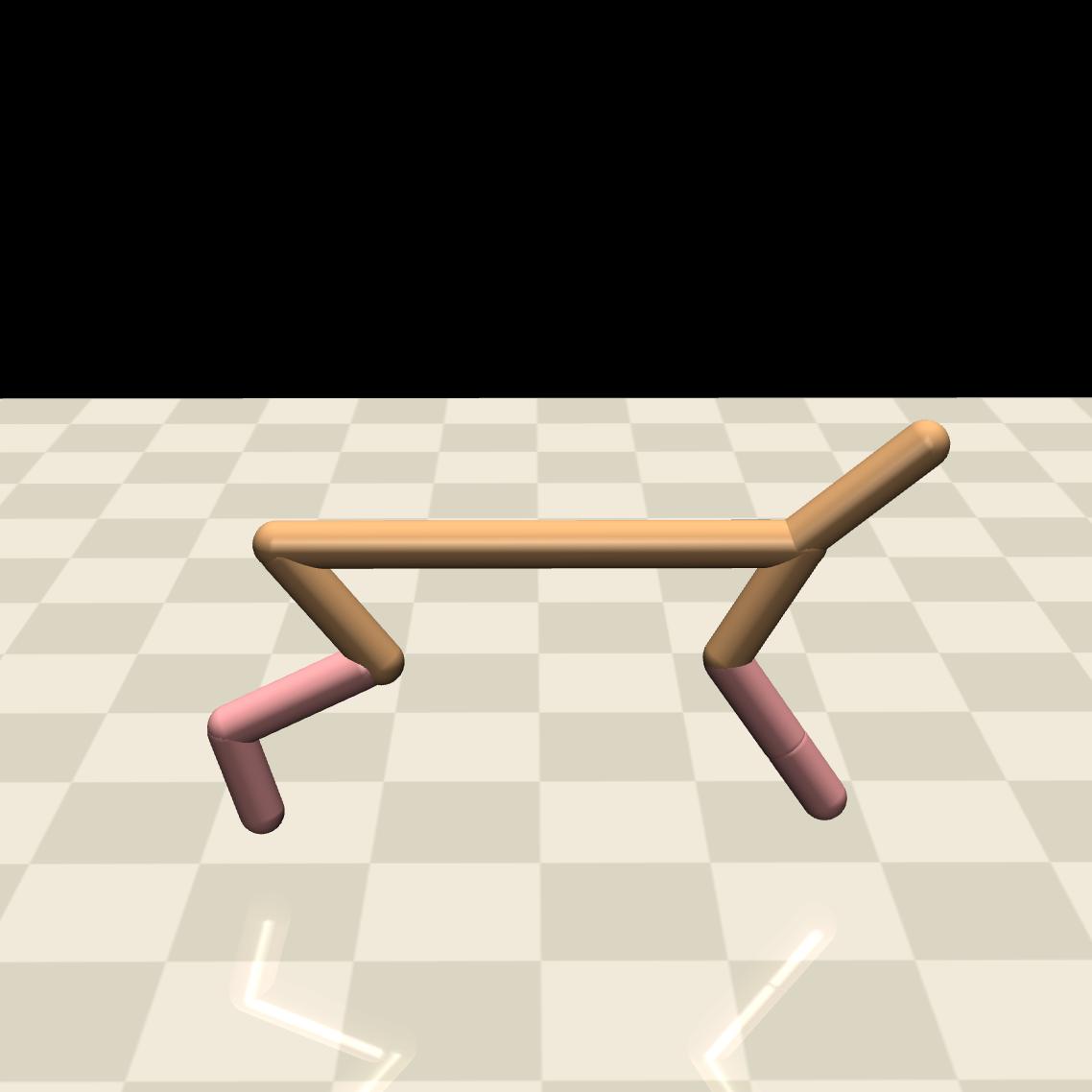}
        \smallskip
        \textbf{(c)} Half Cheetah
    \end{minipage}
\end{minipage}
\caption{Different robots used within the tasks of the evaluated test-suites.}
\label{fig:appendix-robots}
\end{figure}

\begin{itemize}[noitemsep, nolistsep]
    \item \textbf{Car robot:} This robot has two wheels on the rear that the agent can control with one free-rolling front wheel. Steering and movement require nuanced coordination. The action space for the car is $\left[-1, 1\right]^{2}$, and the agent is shown in Figure~\ref{fig:appendix-robots}a.
    \item \textbf{Fetch Manipulator:} A 7-DoF arm with a two-finger parallel gripper, controlled via small end-effector displacements in Cartesian space, with inverse kinematics handled internally. The action space is $\left[-1, 1\right]^{4}$, and the agent is shown in Figure~\ref{fig:appendix-robots}b.
    \item \textbf{HalfCheetah:} This robot is a planar bipedal agent with six actuated hinge joints. The action space is $\left[-1, 1\right]^{6}$, with each dimension applying torque to one joint. The agent is shown in Figure~\ref{fig:appendix-robots}c.
    \item \textbf{CarGoal1}: In this environment, the car-robot must steer to a goal region while avoiding nine hazards whose locations are randomised each episode. The agent, the goal, and hazards are placed randomly (within bounds) at reset. At each timestep, the agent observes LIDAR readings to the goal and to hazards (each with separate lidar observations), along with its own pose and velocity (forming a vector observation). The reward is composed of a “distance” term that gives $r_t = (D_{\text{last}} - D_{\text{now}})\,\beta$, plus a bonus $R_{\mathrm{goal}}$ for reaching the goal. Costs are incurred when the agent comes into contact with any hazard (each hazard contact incurs a cost of 0.2). Episodes terminate after reaching the goal or exceeding a time limit (1000 steps).  
    \item \textbf{CarGoal2}: This environment extends CarGoal1 by increasing the number of hazards to twenty, all with randomised positions at each episode reset. As in CarGoal1, the car-robot, the goal, and the hazards are placed randomly (within bounds). The agent receives observations consisting of separate LIDAR readings to the goal and to hazards, together with its own pose and velocity. The reward function is identical, with a distance-based shaping term $r_t = (D_{\text{last}} - D_{\text{now}})\,\beta$ and a goal-reaching bonus $R_{\mathrm{goal}}$. Costs are incurred whenever the agent makes contact with hazards (each hazard contact incurs a cost of 0.2). Episodes terminate either upon reaching the goal or after 1000 steps. 
    \item \textbf{CarButton1}: In this environment, the car-robot must drive to and press a correct button while avoiding nine hazards (including three incorrect buttons) whose positions are randomised each episode. At reset, the agent, buttons, and hazards are randomly placed within bounds. At each timestep, the agent observes separate LIDAR readings to the goal, buttons and to hazards, as well as its own pose and velocity (vector observation). The reward consists of a shaping term \(r_t = (D_{\text{last}} - D_{\text{now}})\,\beta\) plus a bonus \(R_{\mathrm{button}}\) for successfully pressing the correct button. A cost (e.g.\ 0.2) is incurred whenever the agent contacts any hazard or an incorrect button. Episodes terminate upon pressing the button or exceeding a timestep limit (e.g.\ 1000 steps).
    \item \textbf{CarButton2}: This environment extends CarButton1 by increasing the number of hazards to fourteen (including three incorrect buttons), all with randomised positions at each episode reset. As in CarButton1, the car-robot, buttons, and hazards are placed randomly within bounds. The agent observes separate LIDAR readings to the goal, to buttons, and to hazards, together with its own pose and velocity. The reward function is identical, with a shaping term \(r_t = (D_{\text{last}} - D_{\text{now}})\,\beta\) and a bonus \(R_{\mathrm{button}}\) for pressing the correct button. A cost (0.2) is incurred whenever the agent contacts any hazard or an incorrect button. Episodes terminate either upon pressing the correct button or after 1000 steps.
    \item \textbf{FetchReach}: In this environment, a 7-DoF Fetch manipulator robot must move its end effector to a randomly selected 3D goal position and maintain it. At reset, the desired goal is sampled uniformly within the reachable workspace. At each timestep, the agent receives a reduced observation consisting of the end effector position and velocity (6 dimensions), concatenated with the 3D goal position. The action space is $\mathrm{Box}(-1,1)^3$, representing Cartesian displacements $(dx, dy, dz)$ of the end effector (the gripper action is fixed to zero in this task). The reward is the negative Euclidean distance between the achieved and desired goal, with an additional bonus of $+5.0$ when the goal is reached within a tolerance of 0.05\,m. Safety costs are imposed through two terms: (1) a velocity cost of $1.0$ whenever the magnitude of the end effector velocity exceeds a threshold $v_{\max} = 0.01$, and (2) a jerk penalty $\lambda_{\mathrm{jerk}} \sum_{i} (\Delta v_i)^2$ with $\lambda_{\mathrm{jerk}} = 1000.0$. The total episodic cost is the sum of these penalties across the trajectory. Episodes terminate upon reaching the goal or after a maximum horizon of 500 steps.
    \item \textbf{HalfCheetah}: The HalfCheetah robot is a planar serial-link agent with six actuated hinge joints, controlled by torques in the action space $\mathrm{Box}(-1,1)^6$. The observation is the 17-dimensional vector of generalised positions (excluding root \(x\)) and velocities. The nominal reward is the forward velocity (distance travelled per timestep). In the Safe Velocity variant, a safety cost of $1.0$ is incurred on any timestep where the magnitude of the agent’s velocity exceeds a predefined threshold (set to 50\% of its maximum speed under converged PPO). Episodic length is 500 steps.
\end{itemize}

\begin{figure}[tb]
\centering
\begin{minipage}{0.95\textwidth}
    \begin{minipage}[t]{0.15\linewidth}
        \centering
        \includegraphics[width=\linewidth]{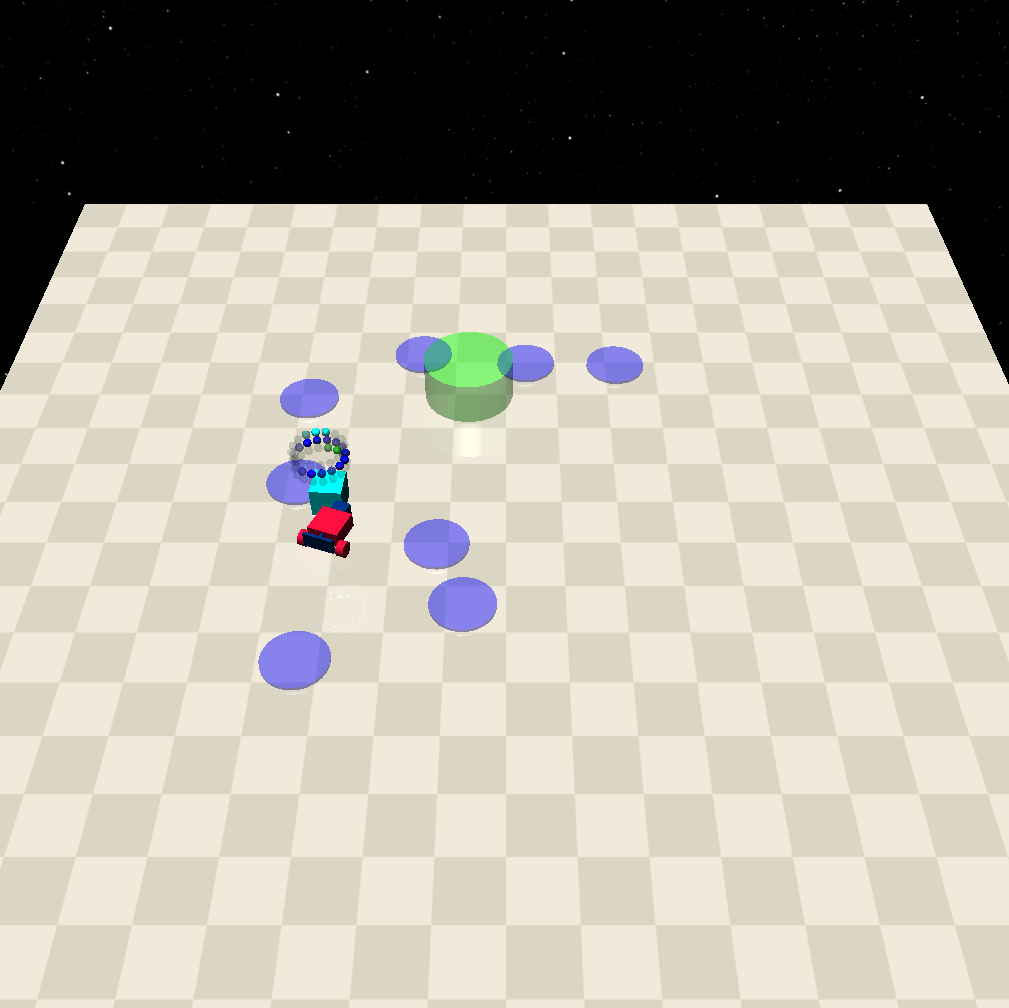}
        \smallskip
        \textbf{(a)} CarGoal1
    \end{minipage}
    \hfill
    \begin{minipage}[t]{0.15\linewidth}
        \centering
        \includegraphics[width=\linewidth]{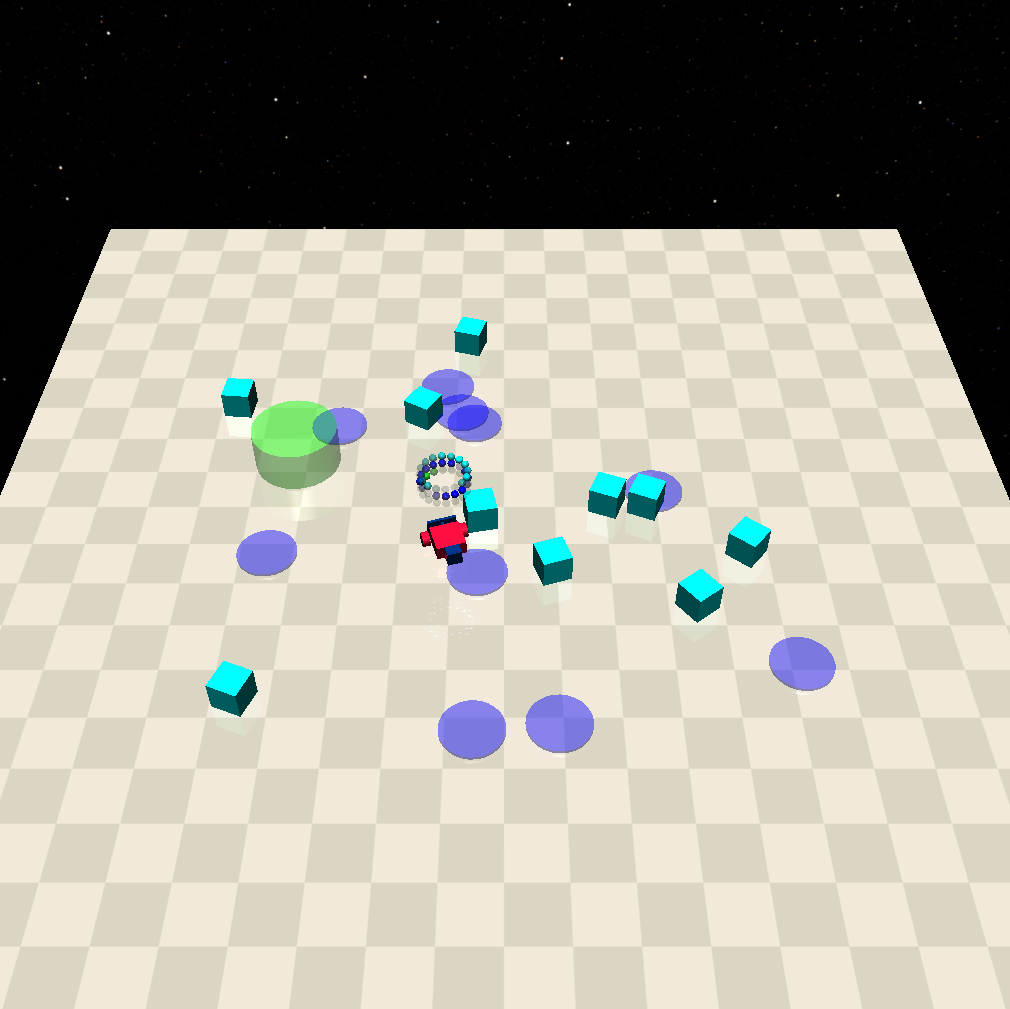}
        \smallskip
        \textbf{(b)} CarGoal2
    \end{minipage}
    \hfill
    \begin{minipage}[t]{0.15\linewidth}
        \centering
        \includegraphics[width=\linewidth]{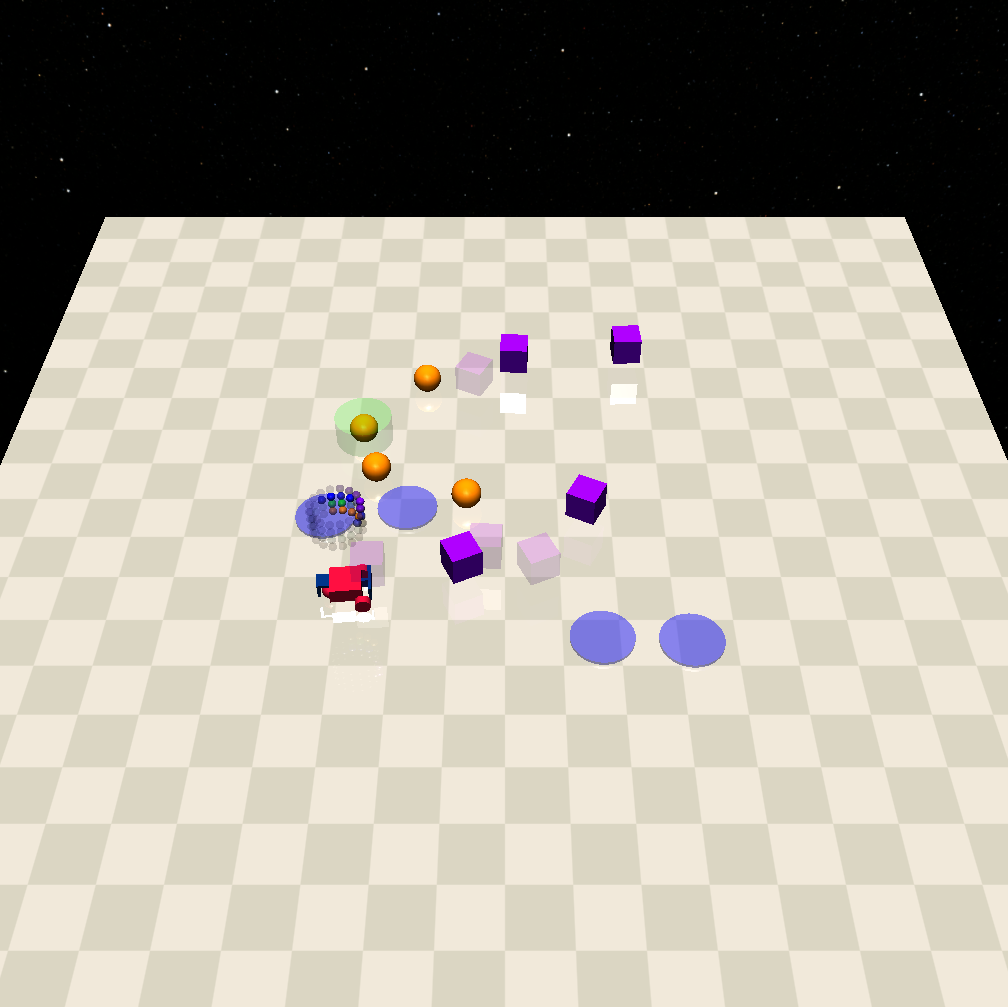}
        \smallskip
        \textbf{(c)} CarButton1
    \end{minipage}
    \hfill
    \begin{minipage}[t]{0.15\linewidth}
        \centering
        \includegraphics[width=\linewidth]{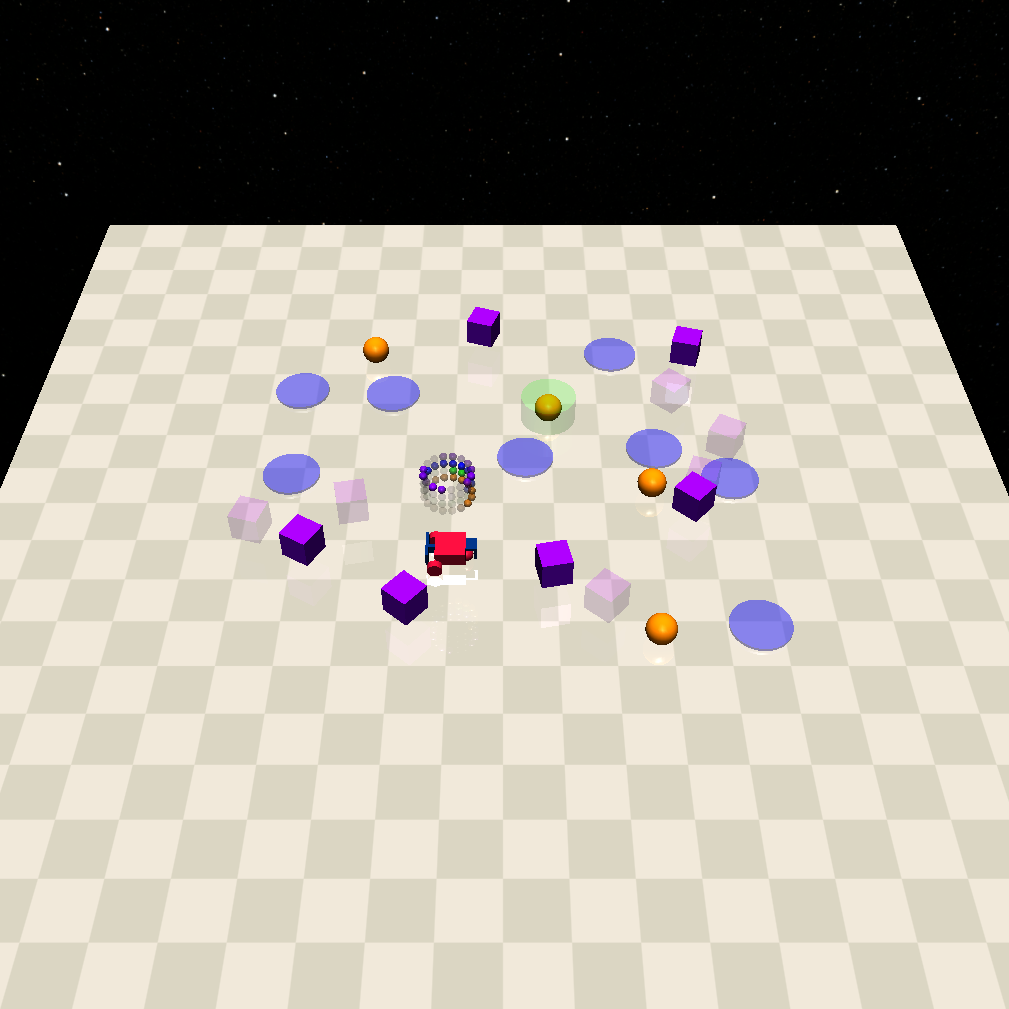}
        \smallskip
        \textbf{(d)} CarButton2
    \end{minipage}
    \hfill
    \begin{minipage}[t]{0.15\linewidth}
        \centering
        \includegraphics[width=\linewidth]{Images/Appendix/fetch.png}
        \smallskip
        \textbf{(e)} FetchReach
    \end{minipage}
    \hfill
    \begin{minipage}[t]{0.15\linewidth}
        \centering
        \includegraphics[width=\linewidth]{Images/Appendix/cheetah.jpeg}
        \smallskip
        \textbf{(f)} Half Cheetah
    \end{minipage}
\end{minipage}
\caption{Example navigation tasks with varying complexity levels for evaluating USC.}
\label{fig:appendix-task-vis}
\end{figure}

\FloatBarrier
\endgroup


\section{Theoretical Analysis}
\label{sec:appendix-theo-analysis}
In this section, we theoretically analyse USC and show that USC is more conservative than other alternatives under certain conditions.

\subsection{Theorem~\ref{thm:1}}

\textbf{Theorem 1.} 
The log-exp-sum (LSE) penalty, used in Equation~\ref{eq:safety-critic-loss}, is more effective at encouraging conservatism when the $m$ alternative actions are sampled uniformly over the action space, rather than from the current policy~\cite{bharadhwaj2020conservative}. If the policy aims to avoid high-cost regions, then the uniform sampling is more likely to draw from high-cost out-of-distribution actions. Formally, for any fixed state $s$, the conservative term is strictly larger under uniform scaling.

\begin{proof}
For a fixed state $s \in S$, and the action space $A$, the safety critic satisfies:
\begin{equation}
L(s) \leq Q_C(s,a) \leq U(s) \quad \forall a \in A
\end{equation}
where $L(\cdot)$ and $U(\cdot)$ are the lower and upper bounds of the safety critic's output at state $s$. We let $\tau(s) \in (L(s), U(s))$ be a cost threshold and:
\begin{equation}
\text{Risk}_{\tau(s)} := \{ a \in A | Q_C(s,a) \geq \tau(s) \}
\end{equation}
be the most costly set of actions above the cost threshold $\tau(s)$. For a distribution $P$ over actions (either $\pi(\cdot|s)$ or uniform), the probability of hitting that set is:
\begin{equation}
p_P(s) := \mathbb{P}_{A\sim P}(A \in \text{Risk}_{\tau(s)}) = \mathbb{P}(Q_C(s,a) \geq \tau(s)))
\end{equation}
We assume the standard safe RL condition: the policy is less likely to select risky actions than the uniform sampler:
\begin{equation}
p_\pi(s) \leq p_{\text{unif}}(s) - \eta_s
\end{equation}
for some $\eta_s > 0$. This assumption reflects the idea that the agent's current policy tends to avoid high-cost actions, whereas the uniform sampler does not. Let $a_1, \ldots, a_m \sim P$ be independent and identically distributed $m$ sampled actions. For shared scalars, temperature $c(s) > 0$ and batch mean $\mu(s)$, the normalised critic value for each sampled action and the LSE is defined as:
\begin{equation}
\begin{aligned}
z_i := c(s)(Q_C(s,a_i) - \mu(s)) \\
\text{LSE}^{(P,c)}_m(s) := log \sum^m_{i=1}e^{z_i}
\end{aligned}
\end{equation}
We define the conservative margin at state $s$ under sampler $P$ as:
\begin{equation}
\Delta^{(c)}_p(s):= \text{LSE}^{(P,c)}_m(s) - c(s)(Q_C(s, a_{\text{real}}) - \mu(s))
\end{equation}
where $a_{\text{real}}$ is the ground truth action from the data. This gives us the gap between the sampled actions and the ground truth under any sampler $P$. Since the data (second term) term is identical under both samplers, $P \in \{ \pi(\cdot|s), \text{Unif} \}$, the difference $\Delta^{(c)}_\text{Unif} - \Delta^{(c)}_\pi$ reduces to comparing the two LSE terms.

To perform this comparison, we recall that LSE is a smooth version of the maximum: it always lies between the hard max and the hard max plus $\log m$. Evidently, if we can show the expected hard max of $Q_C$ values is larger under uniform sampling, then the same must hold for the LSE- up to $\log m$.

As stated, LSE acts like a smoothed max over $m$ inputs. This is formalised by the bound:
\begin{equation}
\max_i z_i \leq \log \sum^m_{i=1} e^{z_i} \leq \max_i z_i + \log m
\end{equation}
So, this shows that LSE differs from the true max by at most $\log m$. So, if the expected max of $z_i$ is larger under uniform sampling, then the expected LSE will also be larger up to $\log m$. Applying this logic to both samplers gives:
\begin{equation}
\mathbb{E}[\text{LSE}^{(\text{Unif},c)}_m(s)] - \mathbb{E}[\text{LSE}^{(\pi,c)}_m(s)] \geq c(s)(\mathbb{E}[\max_i Q_C(s, a_i)]_{\text{Unif}} - \mathbb{E}[\max_i Q_C(s, a_i)]_{\pi}) - \log m
\end{equation}
To show that the expected max of a uniform sampling is larger, we need to find the lower bound of $\mathbb{E}[\max_i Q_C(s, a_i)]$ for either sampler $P$. Let:
\begin{equation}
K_P := \sum^m_{i=1} \mathbf{1}\{ a_i \in \text{Risk}_{\tau(s)} \}
\end{equation}
be a Binomial random variable counting how many of the sampled actions fall into the risky set. Thus,
\begin{equation}
\mathbb{P}(\max_i Q_C(s, a_i) \geq \tau(s)) = \mathbb{P}(K_P \geq 1) = 1-(1-p_P(s))^m
\end{equation}
is the probability that at least one sample is risky. On the event that $K_P = 0$, the max is at least $L(s)$ as all sampled values are below $\tau(s)$, but the safety critic is still bounded by $L(s)$. Therefore:
\begin{equation}
\mathbb{E}[\max_iQ_C(s,a_i)]_P \geq L(s) + (\tau(s) - L(s))(1-(1-p_P(s))^m)
\end{equation}
This shows that the expected maximum depends only on the probability of sampling a risky action. This, whichever sampler assigns more probability mass to the risky set $\text{Risk}_{\tau(s)}$ will yield the larger expected maximum. By assumption, $p_{\text{Unif}}(s) > p_\pi(s)$, so:
\begin{equation}
(1- p_\pi(s))^m > (1-p_{\text{Unif}}(s))^m
\end{equation}
Therefore, the uniform sampler is more likely to include at least one risky action, leading to a larger expected maximum:
\begin{equation}
\mathbb{E}[\max Q_C]_{\text{Unif}} - \mathbb{E}[\max Q_C]_{\pi} \geq (\tau(s) - L(s))[(1- p_\pi(s))^m - (1-p_{\text{Unif}}(s))^m]
\end{equation}
Plugging this into the LSE inequality:
\begin{equation}
\mathbb{E}[\text{LSE}^{(\text{Unif}, c)}_m(s) - \text{LSE}^{(\pi, c)}_m(s)] \geq c(s)(\tau(s) - L(s))[(1- p_\pi(s))^m - (1-p_{\text{Unif}}(s))^m] - \log m
\end{equation}
As shown earlier, because both gaps $\Delta^{(c)}_P(s)$ share the same data term, we conclude:
\begin{equation}
\mathbb{E}[\Delta^{(c)}_{\text{Unif}}(s) - \Delta^{(c)}_{\pi}(s)] \geq c(s)(\tau(s) - L(s))[(1- p_\pi(s))^m - (1-p_{\text{Unif}}(s))^m] - \log m
\end{equation}
This shows us that, in expectation, drawing actions uniformly instead of from the policy always gives a larger conservative penalty. The margin is proportional to: how high the risky region is above the baseline $(\tau(s) - L(s))$, and how many risky hits you expect to get under uniform sampling versus policy sampling.
\end{proof}

\paragraph{Remark (Effect on the loss).}
From the theorem, for any fixed state $s$ and sample size $m \geq 1$:
\begin{equation}
\mathbb{E}\!\left[\Delta^{(c)}_{\mathrm{Unif}}(s)-\Delta^{(c)}_{\pi}(s)\right]
\ge
c(s)(\tau(s)-L(s))\big((1-p_\pi(s))^m-(1-p_{\mathrm{Unif}}(s))^m\big)-\log m.
\label{eq:unif-vs-pi-gap}
\end{equation}
Hence, whenever the right-hand side of Equation~\ref{eq:unif-vs-pi-gap} is positive, the conservative penalty term inside Equation~\ref{eq:safety-critic-loss} is, in expectation, strictly larger under uniform sampling than under policy sampling:
\begin{equation}
\mathbb{E}\!\left[\Delta^{(c)}_{\mathrm{Unif}}(s)\right]
\geq
\mathbb{E}\!\left[\Delta^{(c)}_{\pi}(s)\right]
+ c(s)(\tau(s)-L(s))\big((1-p_\pi(s))^m-(1-p_{\mathrm{Unif}}(s))^m\big)-\log m.
\end{equation}
When the uncertainty modulation $\tilde u(s,a_{\text{data}})\ge 0$ is applied as in Equation~\ref{eq:safety-critic-loss}, $L_C^{(c)}(s,a_{\text{data}})=\tilde u(s,a_{\text{data}})\,\Delta^{(c)}(s)$, the inequality is preserved (it is multiplied by the same nonnegative factor on both sides):
\begin{equation}
\mathbb{E}\!\left[L_C^{(c)}\big|\,\mathrm{Unif}\right]
-\mathbb{E}\!\left[L_C^{(c)}\big|\,\pi\right]
\ge
\tilde u(s,a_{\text{data}})\Big(c(s)(\tau(s)-L(s))\big((1-p_\pi(s))^m-(1-p_{\mathrm{Unif}}(s))^m\big)-\log m\Big).
\end{equation}
Consequently, if $\tilde{u}(s,a) > 1$, then the conservative bias is \emph{amplified}, and uniform sampling yields strictly stronger pessimism than policy sampling. If $\tilde{u}(s,a) < 1$, the conservative bias is weakened, but the loss remains more conservative than under policy sampling alone, only scaled down by $\tilde{u}(s,a)$. This scaling typically results in less conservatism than the policy sampling without uncertainty modulation, thus being less conservative. Evidently, the loss will create sharper gradients than alternative methods due to increased conservatism in high-cost, high-uncertainty regions and reduced conservatism in low-uncertainty regions.


\section{Full Training Results}
\label{sec:appendix-full-results}

\subsection{Extended Analysis of Core Results}
\label{sec:appendix-Extended Analysis}

\begin{figure*}[tb]
\centering
    \begin{subfigure}{\textwidth}
    \includegraphics[width=\linewidth]{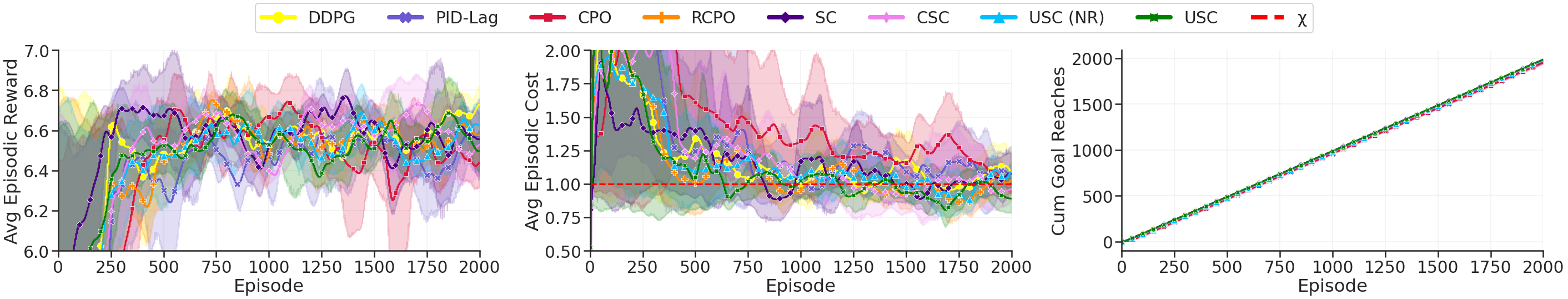}
    \vspace*{-5mm}
    \caption{CarGoal1 environment results}
    \label{fig:core-cargoal1}
    \end{subfigure}
    
    \begin{subfigure}{\textwidth}
    \includegraphics[width=\linewidth]{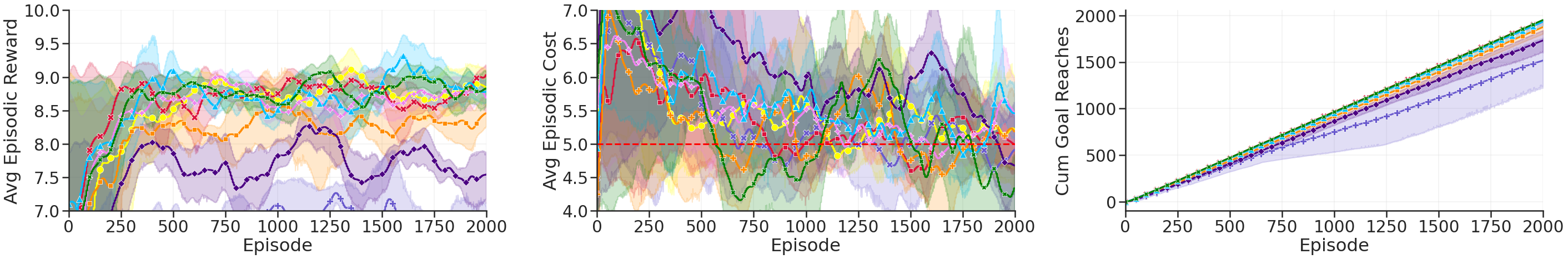}
    \vspace*{-5mm}
    \caption{CarGoal2 environment results}
    \label{fig:core-cargoal2}
    \end{subfigure}

    \begin{subfigure}{\textwidth}
    \includegraphics[width=\linewidth]{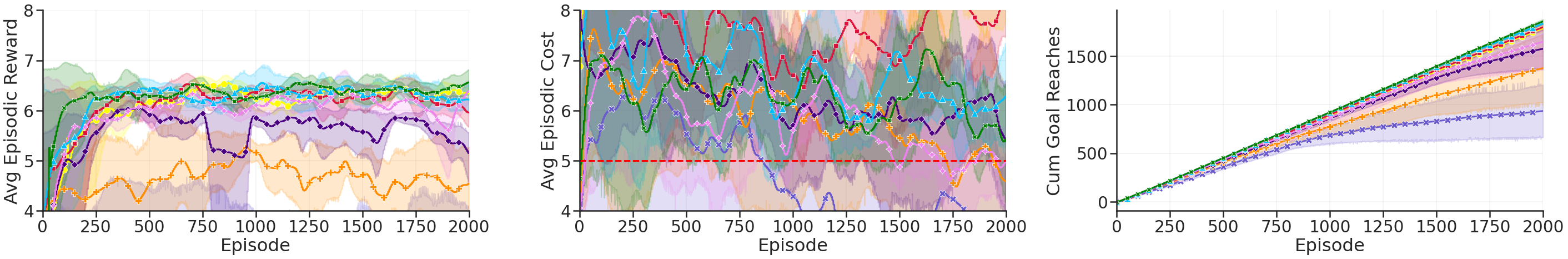}
    \vspace*{-5mm}
    \caption{CarButton1 environment results}
    \label{fig:core-carbutton1}
    \end{subfigure}

    \begin{subfigure}{\textwidth}
    \includegraphics[width=\linewidth]{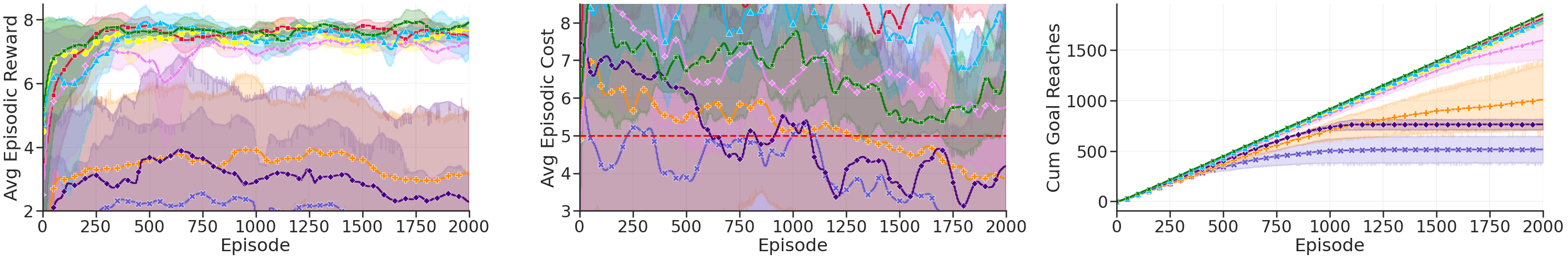}
    \vspace*{-5mm}
    \caption{CarButton2 environment results}
    \label{fig:core-carbutton2}
    \end{subfigure}
\caption{Average episodic reward, cost, and cumulative goal reaches of examined methods (DDPG, Safety Critic, Conservative Safety Critic, Uncertain Safety Critic) for the CarGoal1 (a), CarGoal2 (b), CarButton1 (c), and CarButton2 (d) environments.}
\label{fig:training-results}
\end{figure*}

In Section~\ref{sec:core-results}, we show the main results for all approaches in a set of tasks. Below, in Figure~\ref{fig:training-results}, we show the full training plots for all results/methods shown in Table~\ref{tab:main-results}.

Across all four Safety Gymnasium tasks, the training curves show that USC delivers the sharpest reward–cost trade-off and the most stable learning dynamics. In CarGoal1 and CarGoal2, USC maintains high average reward while keeping average cost close to (or below) the competing safety critic baselines, with visibly tighter confidence bands than SC and CSC. This mirrors the summary statistics in Table~\ref{tab:main-results}, where USC attains the top reward in CarGoal2 ($ \approx 8.84$) while achieving the lowest cost among the learning methods ($ \approx 5.05$ vs. $ \approx 5.36$ for SC), indicating that the improved critic gradients translate directly into better online learning traces.

\begin{figure*}[tb]
\centering
    \begin{subfigure}{\textwidth}
    \includegraphics[width=\linewidth]{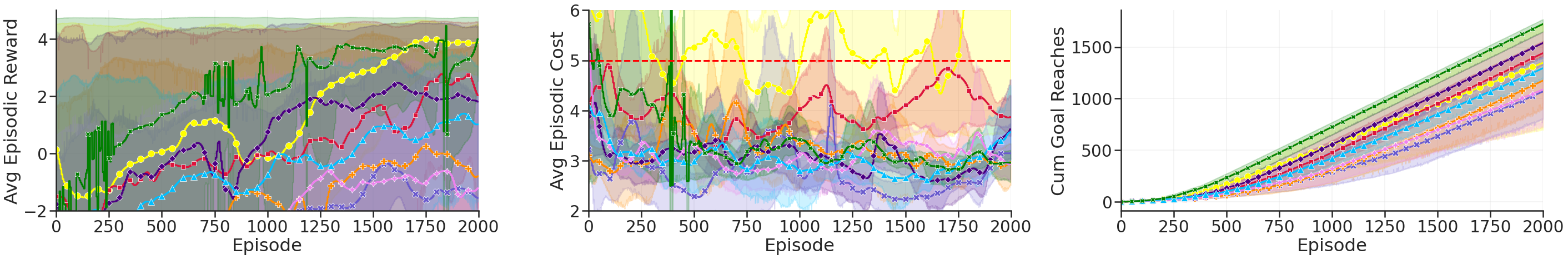}
    \vspace*{-5mm}
    \caption{FetchReach environment results}
    \label{fig:core-fetchreach}
    \end{subfigure}

    \begin{subfigure}{0.66\textwidth}
    \includegraphics[width=\linewidth]{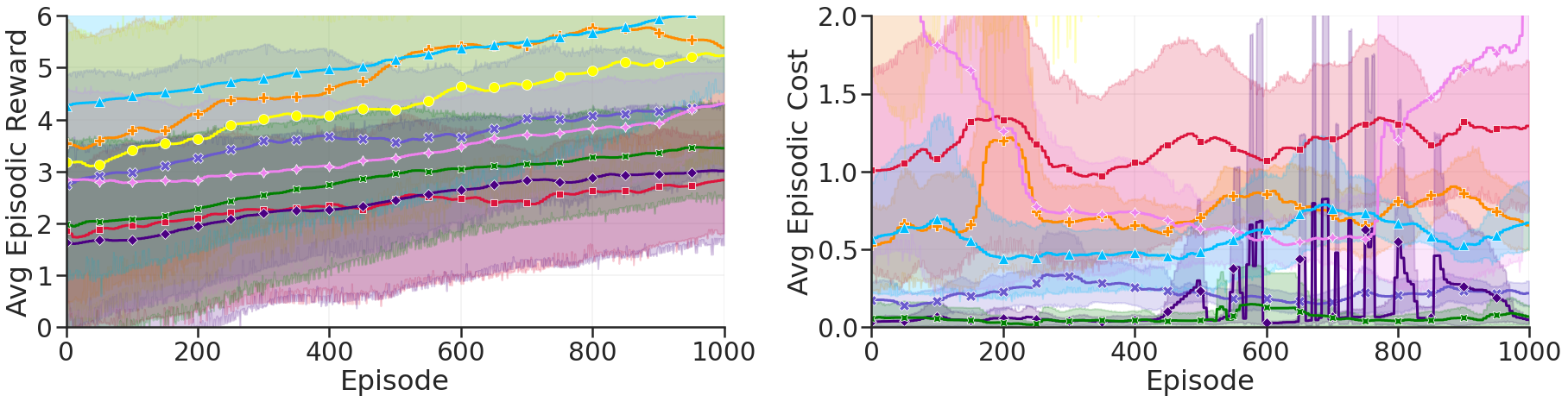}
    \vspace*{-5mm}
    \caption{Mujoco Cheetah environment results}
    \label{fig:core-cheetah}
    \end{subfigure}

\caption{Average episodic reward, average episodic cost, and cumulative goal reaches of examined methods (DDPG, Safety Critic, Conservative Safety Critic, Uncertain Safety Critic) for the FetchReach (a), and HalfCheetah (b) environments.}
\label{fig:training-results-2}
\end{figure*}

As task difficulty scales (CarGoal $\to$ CarButton), the limitations of prior critics become pronounced. The standard Safety Critic suffers reward collapse and erratic cost spikes; in the cumulative goal-reach plots, its curve frequently flattens: evidence that the policy stops solving episodes for extended stretches. In CarButton2, this stagnation is particularly evident: SC’s cumulative successes plateau early ($\approx$ episode 1000), whereas USC continues to accrue goals nearly linearly through training, indicating sustained policy improvement rather than oscillation. The Conservative Safety Critic agent remains more stable than SC but still degrades at larger scales: it avoids the worst collapses yet learns noticeably slower and accumulates substantially fewer successes (as seen at $\approx$ episode 1750 for both CarButton environments), consistent with its over-conservatism and the reward underperformance reported in Table 1 (e.g., CarButton2 reward $\approx$ 6.45 for CSC vs. $\approx$ 7.69 for USC). 

In FetchReach, seen in Figure~\ref{fig:training-results-2}, USC stabilizes around the highest reward levels while maintaining the lowest episodic costs, yielding the steepest cumulative goal-reach curve among all methods. Safety Critic shows highly erratic cost spikes and inconsistent reward, whereas CSC avoids catastrophic collapse but underperforms in both reward and goal completion compared to USC. In fact, USC achieves the highest goal success rate in the environment. These dynamics confirm that USC can scale beyond Safety Gym to more structured manipulation tasks without losing stability.

In HalfCheetah, the divergence between methods is even more pronounced. SC rapidly destabilizes, with its cost curve oscillating wildly, while CSC remains overly conservative and actually never is able to stabilise costs. In contrast, USC achieves monotonic reward growth coupled with suppressed cost variance, demonstrating both efficient exploration and safe policy improvement.

\begin{figure*}[tb]
\centering
    \begin{subfigure}{\textwidth}
    \includegraphics[width=\linewidth]{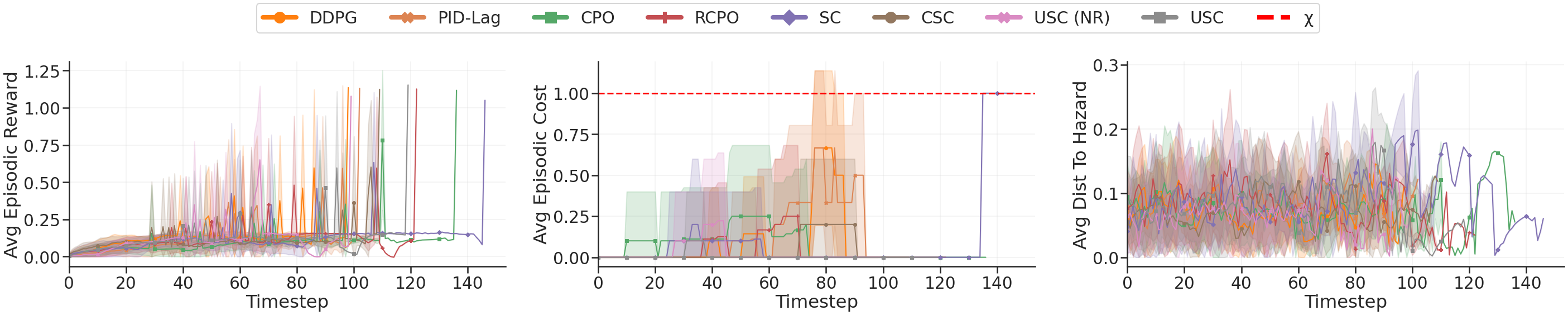}
    \vspace*{-5mm}
    \caption{CarGoal1 environment results}
    \label{fig:test-cargoal1}
    \end{subfigure}
    
    \begin{subfigure}{\textwidth}
    \includegraphics[width=\linewidth]{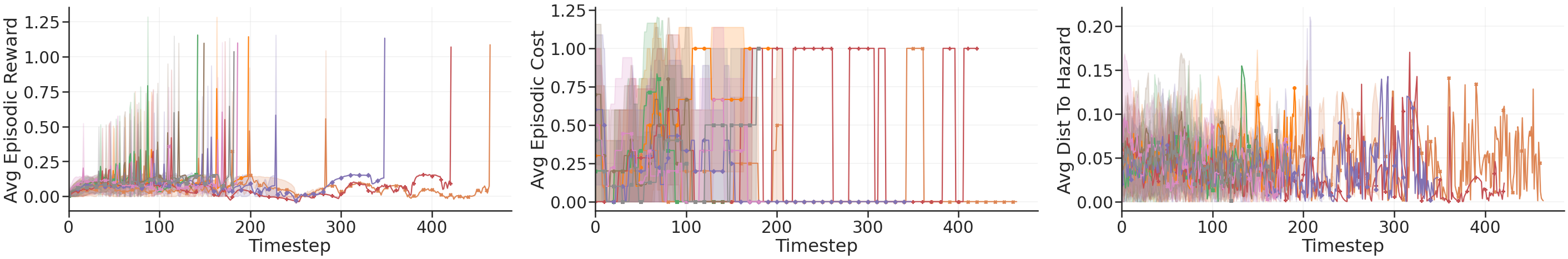}
    \vspace*{-5mm}
    \caption{CarGoal2 environment results}
    \label{fig:test-cargoal2}
    \end{subfigure}

    \begin{subfigure}{\textwidth}
    \includegraphics[width=\linewidth]{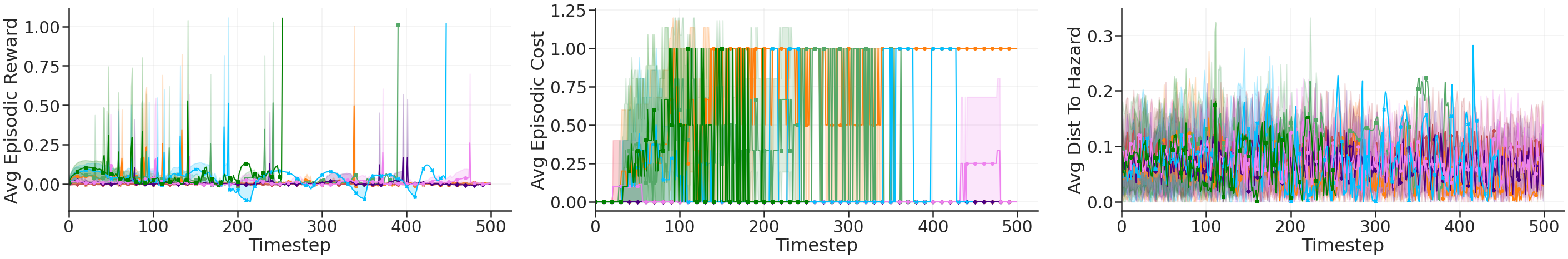}
    \vspace*{-5mm}
    \caption{CarButton1 environment results}
    \label{fig:test-carbutton1}
    \end{subfigure}

    \begin{subfigure}{\textwidth}
    \includegraphics[width=\linewidth]{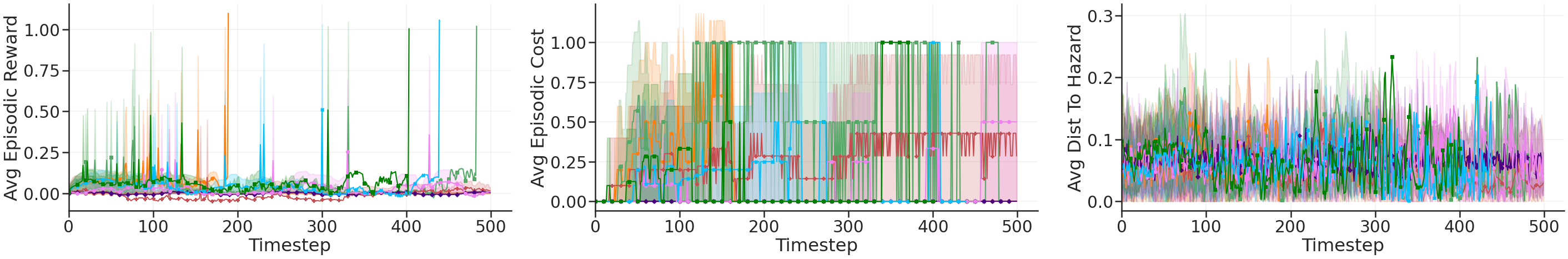}
    \vspace*{-5mm}
    \caption{CarButton2 environment results}
    \label{fig:test-carbutton2}
    \end{subfigure}

\caption{Average episodic reward, average episodic cost, and distance to nearest hazard of examined methods (DDPG, Safety Critic, Conservative Safety Critic, Uncertain Safety Critic) for the CarGoal1 (a), CarGoal2 (b), CarButton1 (c), and CarButton2 (d) environments during 10 test/deployment episodes.}
\label{fig:testing-results}
\end{figure*}

The deployment-phase evaluations, see in Figure~\ref{fig:testing-results}, confirm the trends reported in Table~\ref{tab:usc-deployment-results}. Across all CarGoal and CarButton tasks, USC consistently delivers higher reward while keeping costs relatively low compared to all other methods, and it maintains the highest average distance to hazards. Safety Critic again proves brittle, with frequent cost spikes and poor hazard avoidance, while CSC is more stable but overly conservative, often suppressing reward. Notably, in the more difficult CarButton tasks, USC is the only safety-aware method that sustains balanced performance without collapsing into unsafe or overly cautious behaviour and achieves 100\% success rate, whereas the other safety-aware methods fail to reach the goal every time. These results reinforce that USC’s uncertainty-modulated training advantage translates into reliable and safe deployment performance.

\begin{figure*}[tb]
\centering
    \begin{subfigure}{0.66\textwidth}
    \includegraphics[width=\linewidth]{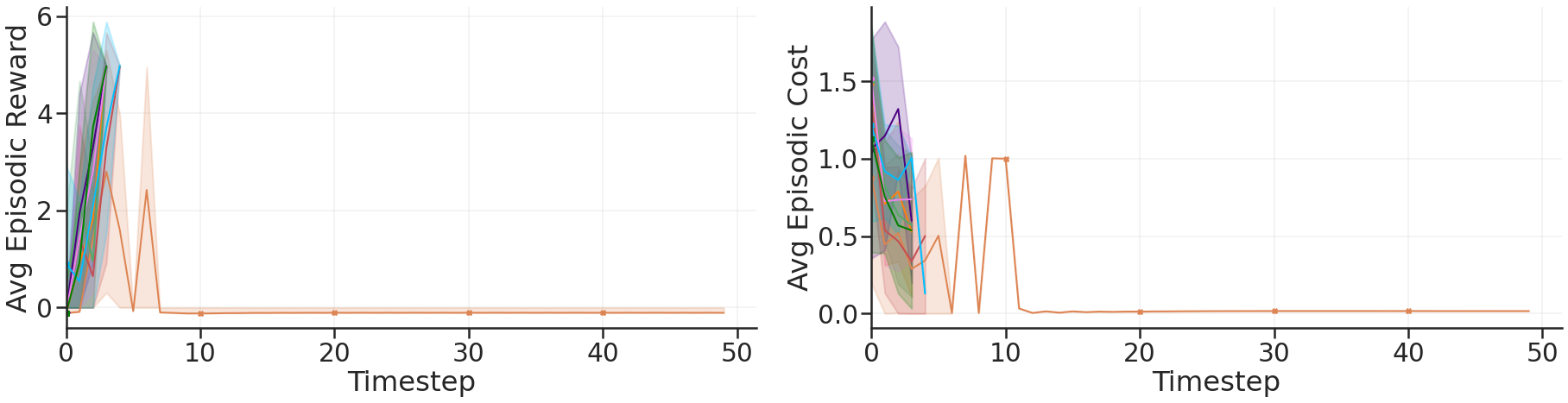}
    \vspace*{-5mm}
    \caption{FetchReach environment results}
    \label{fig:test-fetch}
    \end{subfigure}
    
    \begin{subfigure}{0.66\textwidth}
    \includegraphics[width=\linewidth]{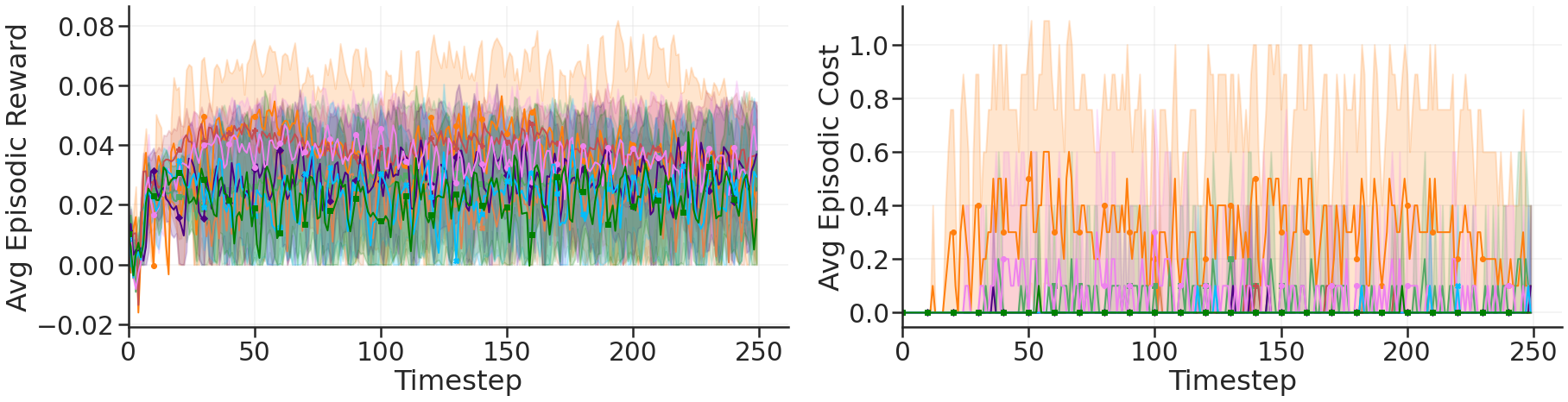}
    \vspace*{-5mm}
    \caption{HalfCheetah environment results}
    \label{fig:test-cheetah}
    \end{subfigure}

\caption{Average episodic reward, average episodic cost, and distance to nearest hazard of examined methods (DDPG, Safety Critic, Conservative Safety Critic, Uncertain Safety Critic) for the FetchReach (a), and HalfCheetah (b) environments during 10 test/deployment episodes.}
\label{fig:testing-results-2}
\end{figure*}

The deployment results in FetchReach further highlight USC’s balanced behaviour: it achieves the strongest reward growth while driving down episodic cost over time, outperforming SC and CSC which either fluctuate or stall in reward–cost trade-offs. In HalfCheetah, where instability is common, SC and DDPG show erratic cost spikes and inconsistent reward traces, while CSC remains conservative. By contrast, USC delivers steady low-cost performance with the most reliable hazard avoidance. Together, these results extend our core findings, demonstrating that USC generalises safe and effective deployment across both manipulation and locomotion domains.

\section{Hyperparameter Analysis and Computational Overheads}
\label{sec:appendix-hyperparams}

This section lists the hyperparameters used by all models and USC. Table~\ref{tab:hyperparameter-configuration} summarises all hyperparameters used in Section~\ref{sec:Evaluation}. We will refer the reader to our source code repository for the remaining details.

Using all model configurations in Table~\ref{tab:hyperparameter-configuration}, a single USC run (one random seed) takes 12, 20, 20, 20, 1, and 12 hours of training, respectively, in the CarGoal1, CarGoal2, CarButton1, CarButton2, FetchReach, and HalfCheetah environments. For all other approaches, a single run takes 6, 10, 12, 12, 1, and 10 hours in the same environments, respectively. All experiments were run on a large computing cluster utilising two Nvidia H100 GPUs, 16 CPUs, and up to 500GB memory.

\begin{table}[tb]
\vspace{4mm}
\centering
\captionsetup{justification=centering}
\caption{Summary of hyperparameters in the DDPG algorithm and the ADVICE shield.}
\label{tab:hyperparameter-configuration}
\begin{tabular}{@{}lllll@{}}
\toprule
Parameter & DDPG &  & Parameter & Safety Critics\\ \midrule
Network size & (256, 256) &  & $\delta$& 1e-6\\
Optimizer & Adam &  & $\alpha$& 1e-3\\
Actor learning rate & 2e-3 &  & $\epsilon_{\text{safe}}$& 0.3\\
Critic learning rate & 1e-3 &  & & \\
Size of replay buffer & 2e5 &  & & \\
Batch size & 64 &  & & \\
Gamma & 0.95 &  & & \\
Tau & 5e-3 &  & & \\
Ornstein-Uhlenbeck noise & 0.2 &  & & \\
\bottomrule
\end{tabular}
\end{table}

Hyperparameters for the DDPG algorithm started with author recommendations~\cite{ddpg-paper}. They were manually tuned afterwards to achieve a high performance on individual environments before tests were carried out, meaning the RL algorithm for all approaches was of high performance and fair comparison.


\end{document}